%% file: iclr2027_conference.tex
\documentclass{article} 
\usepackage{iclr2027_conference,times}

\input{math_commands.tex}

\usepackage{booktabs} 
\usepackage{multirow}
\usepackage{caption}
\usepackage{algorithm}
\usepackage{algpseudocode}
\usepackage[dvipsnames]{xcolor}
\usepackage[table]{xcolor}
\usepackage{hyperref}
\hypersetup{
 colorlinks=true,
 citecolor=RoyalBlue,
 linkcolor=BrickRed,
 urlcolor=CornflowerBlue
}
\usepackage{dashrule}
\usepackage{placeins}

\usepackage{makecell}
\usepackage{siunitx}
\usepackage{graphicx}
\usepackage{xurl}
\usepackage{booktabs}
\usepackage{tabularx}
\usepackage{wrapfig}

\newcommand{\ours}{SlotMemory}
\definecolor{catgray}{gray}{0.9}

\title{\ours{}: Object-Centric KV Memory for Streaming Long-Video Generation}

\author{
Weijia Dou\textsuperscript{1} \quad Hui Li\textsuperscript{1} \quad Jiahao Cui\textsuperscript{1} \quad
Lei Zhou\textsuperscript{3} \quad Jingdong Wang\textsuperscript{4} \quad
Siyu Zhu\textsuperscript{1,2,}\thanks{Corresponding author. Email: siyuzhu@fudan.edu.cn}\\
\textsuperscript{1}Fudan University \quad
\textsuperscript{2}Shanghai Innovation Institute \quad
\textsuperscript{3}Meta Superintelligence Labs \quad
\textsuperscript{4}Baidu
}

\iclrfinalcopy 
\begin{document}

\maketitle
\vspace{-26pt}

\begin{tabular}{@{}ll@{}}
\textbf{~~~~Code Repository:} & \url{https://github.com/tj12323/SlotMemory}
\end{tabular}

\input{sec/0_abstract.tex}
\input{sec/1_intro.tex}
\input{sec/2_relatedwork.tex}
\input{sec/3_method.tex}

\input{sec/4_exp.tex}
\input{sec/5_conclusion.tex}

\subsection*{AI use statement}

In this work, we used generative AI tools to improve the readability, grammar, clarity, and overall presentation of the manuscript, and to identify potential typographical, grammatical, and formatting errors. Additionally, generative AI tools were used for language polishing and proofreading of portions of the manuscript. All AI-assisted text was carefully reviewed and revised by the authors to ensure that it accurately reflects the underlying research and does not introduce unsupported claims, factual errors, or changes to the intended technical meaning.

\subsection*{Ethics statement}

This work does not introduce or release a new dataset containing personal or sensitive information. The proposed method is a general-purpose memory mechanism for long-form video generation and is evaluated on synthetic video-generation tasks. We are not aware of any additional ethical, privacy, security, or legal concerns specific to the methodology or experiments presented in this work.

\subsection*{Reproducibility statement}

We provide detailed descriptions of the proposed architecture, memory organization, slot-based KV compression, persistent entity bank, retrieval and update procedures, training objectives, and streaming state transitions in the main paper and appendix. In particular, the appendix specifies the notation, tensor shapes, layer placement, memory budgets, causal update order, entity association and retrieval procedures, training schedule, and optimization settings. The experimental section further documents the Wan2.1-T2V-1.3B backbone, training data, benchmark settings, evaluation protocols, and hardware configuration. 
 To facilitate reproducibility and future research, we plan to release the model and source code of our method publicly after publication, together with the relevant configuration files and implementation details. These materials, combined with the detailed specifications provided in the paper and appendix, are intended to support reproduction of the proposed method and experimental results.



\bibliography{iclr2027_conference}
\bibliographystyle{iclr2027_conference}

\newpage
\clearpage
\input{sec/6_appendix.tex}

\end{document}

%% file: math_commands.tex
\usepackage{amsmath,amsfonts,bm}

\def\Figref#1{Figure~\ref{#1}}

\def\Secref#1{Section~\ref{#1}}

\def\eqref#1{equation~\ref{#1}}

\def\1{\bm{1}}

\DeclareMathAlphabet{\mathsfit}{\encodingdefault}{\sfdefault}{m}{sl}
\SetMathAlphabet{\mathsfit}{bold}{\encodingdefault}{\sfdefault}{bx}{n}



%% file: sec/0_abstract.tex
\begin{abstract}

Streaming video generation models typically rely on temporal-centric memory, organizing historical context as latent queues, compressed context tokens, or unclustered token sets. This often causes identity drift and semantic inconsistency when entities exit the frame or during interactive prompt transitions. To address these limitations, we propose \ours{}, an object-centric Key-Value (KV) memory mechanism for streaming video diffusion. Our approach shifts memory abstraction from ``when'' an event occurred to ``what'' is represented by decomposing the transformer's KV manifold into discrete, reusable semantic slots. Because the slots serve only as routing addresses and the stored content remains the original transformer KV tokens, the mechanism adds entity-level persistence and prompt-aware retrieval across long horizons. Evaluated on 60-second interactive narratives using the Wan2.1-T2V-1.3B backbone, \ours{} achieves the highest Dynamic Degree among evaluated streaming methods and the second-highest aggregate Quality Score, preserving prompt adherence rather than trading it away for consistency. At the 30-second horizon, \ours{} leads in Total, Quality, and Dynamic Degree, improving Dynamic Degree by 31.2\% over MemFlow, the strongest memory-based baseline. Our results support treating structured semantic representation, rather than raw temporal capacity, as a critical primitive for persistent long-form video synthesis.
\end{abstract}

%% file: sec/1_intro.tex
\section{Introduction}
\label{sec:intro}
The frontier of generative video modeling has rapidly evolved from synthesizing isolated, 
short-duration clips~\citep{wan2025wan, hacohen2024ltx} toward the more formidable objective of producing long-form, 
persistent narratives. 
Current streaming frameworks address this challenge by generating video in chunks and conditioning each new chunk on the past, 
either by rolling a window of recent KV states~\citep{yang2025longlive, huang2026self, yesiltepe2025infinity, chen2025skyreels} 
or by injecting a persistent appearance anchor into the generator~\citep{henschel2025streamingt2v}. 
Both families do carry information beyond the recent window---an appearance anchor is a memory that outlives the local slice---but neither addresses that memory by entity: the retained context is not attached to a specific identity, and no mechanism decides when an identity should be kept, recalled, or retired. 
However, these architectures typically rely on a temporal-centric memory organization, 
where history is represented as latent queues~\citep{kim2024fifo}, 
compressed context tokens~\citep{gu2025far}, 
or unclustered token sets~\citep{yin2025slow}.
This approach introduces a fundamental bottleneck: 
by treating memory as a linear sequence of temporal slices rather than a structured representation of the scene, 
the models neglect the semantic hierarchy of the visual field. 
Consequently, they frequently fail to maintain identity coherence when entities exit and re-enter the frame or when the synthesis requires long-range interaction.

This temporal-centric memory organization produces two recurring failure modes in long-horizon video synthesis.
First, it undermines entity persistence;
when a character or object exits the local temporal window and re-enters,
the model often fails to recover its unique attributes,
resulting in visual drift or identity switching.
Second, existing architectures are ill-suited for interactive,
multi-prompt generation.
When a prompt transitions--for example, from \emph{a glaciologist drilling ice} to \emph{an avalanche}, the model must retain persistent entities while discarding transient motion context.
Existing approaches decide what to keep either by recency, through fixed decay over the temporal window,
or by similarity to the current prompt~\citep{ji2025memflow, yang2026anchorforcing, yu2025context},
and lack the structural awareness to distinguish persistent entities from transient motion.
Consequently, they suffer from cumulative loss of coherence,
leading to progressive narrative degradation over extended 60-second rollouts.

\begin{figure*}[t]
\centering
\includegraphics[width=1.0\textwidth]{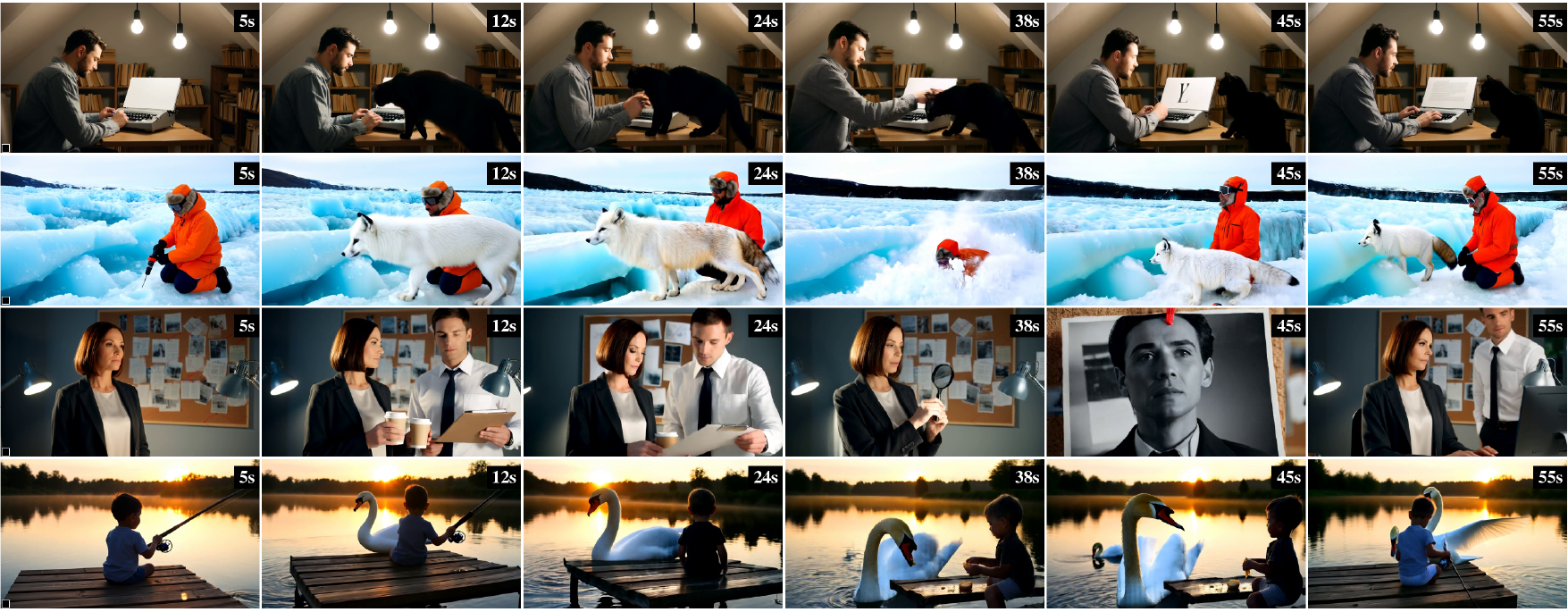}
\vspace{-5mm}
\caption{
\textbf{Teaser examples of persistent, prompt-aware streaming video generation with \ours{}.}
Across extended temporal horizons, \ours{} maintains coherent entity identity and appearance while supporting newly prompted interactions and scene dynamics, highlighting the advantage of object-centric KV memory over purely temporal context organization.
}
\vspace{-5mm}
\label{fig:intro_teaser}
\end{figure*}

To address these limitations, we propose \textbf{\ours{}}, 
an object-centric KV memory mechanism designed to enforce long-term semantic consistency in streaming video diffusion. 
Our core insight is a shift in memory abstraction from temporal occurrence (``when'') to semantic identity (``what''). 
We achieve this by decomposing the transformer's KV manifold into discrete, reusable semantic slots via slot-based representation learning~\citep{locatello2020object, manasyan2025temporally}. 
\Figref{fig:intro_teaser} presents long-form examples where subject identity, key props, and action flow remain coherent across scene changes and temporary occlusions. 
Unlike traditional object-centric models, which are trained for unsupervised discovery on simple scenes~\citep{burgess2019monet, greff2019iodine, kipf2022conditional, elsayed2022savipp,zadaianchuk2023VideoSAUR,manasyan2025temporally}, 
\ours{} bridges structured scene decomposition with high-resolution diffusion. 
Specifically, it utilizes these slots as routing addresses to index and store original, 
transformer-compatible KV tokens. 
This hybrid architecture preserves the high-fidelity detail essential for diffusion while providing a structured, 
entity-level index for efficient retrieval across extended horizons.

Architecturally, \ours{} centers on an iterative read--write--update cycle that maintains a persistent semantic memory bank within fixed computational constraints. 
During the writing phase, 
we leverage a temporally-initialized Slot Attention mechanism~\citep{manasyan2025temporally} to partition high-dimensional transformer manifolds into discrete, 
reusable semantic regions. 
To make the bank usable across prompt switches, 
we recache the text-conditioning cache at every switch while leaving the entity bank and its slot states intact, 
so a new prompt reconditions the rollout without erasing the identities it inherits; 
core entities then survive the transition while obsolete temporal context is evicted. 
By embedding this module directly into the attention stack of the diffusion transformer, 
our architecture enables the model to attend to structured entities rather than unstructured frames, 
fundamentally stabilizing the synthesis of complex, 
multi-stage narratives.

We evaluate \ours{} using the Wan2.1-T2V-1.3B backbone across a comprehensive suite of benchmarks,
including a 60-second interactive multi-prompt scenario.
On the interactive benchmark~\citep{yang2025longlive, ji2025memflow, yesiltepe2025infinity, iamflow, yang2026anchorforcing}, the gains from this shift concentrate in the dimensions that temporal memory degrades first: Dynamic Degree and late-rollout prompt adherence. \ours{} attains the best Dynamic Degree among the evaluated methods and the best text alignment in the final segment (62.97 at 50--60\,s), while its aggregate Quality Score stays within a point of the strongest baseline (81.61 against 82.32). At the 30-second horizon it also leads in Total, Quality, and Dynamic scores, improving Dynamic Degree by 31.2\% over MemFlow, the strongest memory-based baseline.
Across the evaluated horizons, these results support treating structured semantic representation--at least as important as raw temporal capacity--as a critical primitive for persistent and high-fidelity long-form video synthesis.

%% file: sec/2_relatedwork.tex
\section{Related Work}
\label{sec:related}

\noindent\textbf{Long and Streaming Video Generation.}
Autoregressive (AR) streaming frameworks generate chunks sequentially under fixed computational budgets while managing key--value (KV) caches~\citep{kim2024fifo,henschel2025streamingt2v,teng2025magi1}.
Such frameworks supplement local windows with explicit historical memory banks and prompt-aware retrieval~\citep{yang2025longlive,yu2025context}, but these banks store raw frames, chunk segments, or unclustered tokens, so identity is not reusable once an entity leaves the temporal window or the prompt changes.
Retrieval-based designs retain raw fidelity at frame/chunk granularity~\citep{hu2026longliverag,zhao2026omnimem}, while consolidation-based designs compress chunk/segment history into bounded tokens~\citep{lu2026fademem,kim2026memrope}, entangling entity identity with transient dynamics.
Entity-centric designs address reusable entities but fall outside the attention stack: EM-Vid keeps a training-free, entity-indexed bank of sparse latent patches~\citep{vandersanden2026emvid}, and Memento grounds subject memory in reconstruction~\citep{wei2026memento}.
None of the surveyed designs combines entity-level addressing with raw KV fidelity; \ours{} does so by routing original, uncompressed KV tokens through semantic slots in a structured, fixed-budget memory bank inside the diffusion transformer.
A fuller account of this line is given in \Secref{app:related_video}.

\noindent\textbf{Object-Centric Representation Learning.}
Object-centric representation learning decomposes scenes into discrete slots rather than monolithic features~\citep{locatello2020object,kipf2022conditional,wu2023slotformer}.
This line establishes slots as robust persistent primitives, yet treats them as generative latents that decode into pixels or feed downstream predictors, not as transformer-compatible long-term KV memory.
\ours{} bridges the two lines by using slots purely as routing addresses over original KV tokens, imposing object-centric structure on the streaming attention stack without a lossy slot bottleneck.
Further object-centric variants are covered in \Secref{app:related_slots}.

%% file: sec/3_method.tex
\section{Method}
\label{sec:method}
\begin{figure*}[t]
  \centering
  \includegraphics[width=\textwidth]{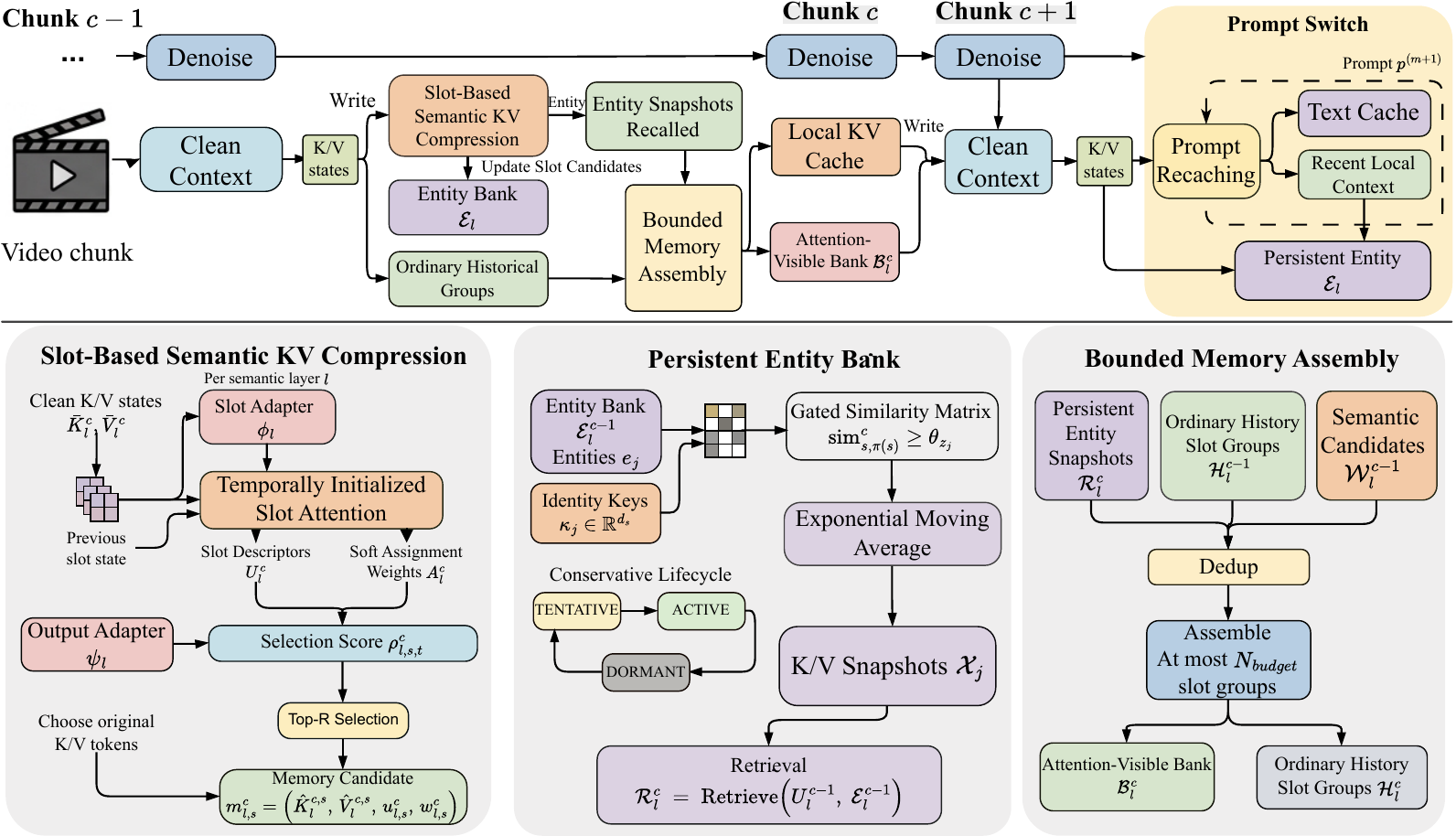}
  \vspace{-5mm}
  \caption{
  \textbf{Overview of \ours{} for streaming long-video generation.}
  Top: each chunk reads the local KV cache and long-term bank, then writes clean states for subsequent chunks.
  Bottom: semantic slots address representative original K/V tokens, a persistent entity bank links identities across chunks and prompt transitions, and bounded assembly selects memory under a fixed budget.
  }
  \vspace{-5mm}
  \label{fig:slotmemory_pipeline}
\end{figure*}

We formulate long-form, multi-prompt video synthesis as bounded-memory streaming generation and propose \ours{}, a layer-wise hybrid memory system for causal video diffusion transformers.
Rather than storing raw frames or unclustered tokens, it retains recent raw KV states for local detail, compresses clean intermediate-layer history into semantically addressable slots, and links those slots to persistent entities that survive temporary absence and prompt changes, so that slots determine \emph{which} past states to recall while the recalled content remains the original K/V consumed by attention.
\Figref{fig:slotmemory_pipeline} shows the resulting read--write--update cycle.

\subsection{Bounded-Memory Streaming: Formulation and Overview}
\label{sec:method_setup}

Given $M$ text prompts $\mathcal{P}=\{p^{(1)},\ldots,p^{(M)}\}$ with switch points $\mathcal{S}=\{s_1,\ldots,s_{M-1}\}$, a causal streaming diffusion transformer synthesizes $C$ consecutive chunks of $T_c$ latent frames, and at switch point $s_m$ the active prompt changes from $p^{(m)}$ to $p^{(m+1)}$ while the accumulated history must remain usable.
At layer $l$ and chunk $c$, the denoising query $Q_l^c$ attends to a short-term local KV cache $\mathcal C_l^c$ augmented by a fixed-capacity long-term KV bank $\mathcal B_l^c$:
\begin{equation}
\label{eq:attn_ctx}
\begin{aligned}
K_l^{\mathrm{ctx},c}&=[K_l^{\mathrm{sink}};K_l^{\mathrm{bank},c};K_l^{\mathrm{local},c}],\\
V_l^{\mathrm{ctx},c}&=[V_l^{\mathrm{sink}};V_l^{\mathrm{bank},c};V_l^{\mathrm{local},c}],\\
Y_l^c&=\operatorname{Attn}(Q_l^c,K_l^{\mathrm{ctx},c},V_l^{\mathrm{ctx},c}).
\end{aligned}
\end{equation}
The sink and local segments come from $\mathcal C_l^c$ and the bank segments from $\mathcal B_l^c$, while cross-attention conditions denoising on a text-conditioning cache refreshed only at switch points.

Two states govern the bank without entering attention directly: the slot state $U_l^c$, which records the slot descriptors of chunk $c$ and initializes the next decomposition, and the persistent entity bank $\mathcal E_l^c$, which tracks visual identities and their K/V snapshots across chunks.
The write path is causally delayed: chunk $c$ reads only states written by earlier chunks, so the states it produces become visible from chunk $c+1$ onward.
After generation, clean-context re-encoding yields $(\bar K_l^c,\bar V_l^c)$, which update the local cache and, at the next write point, are compressed into semantic candidates and committed to the entity bank.
The layer-wise cache layout and state transitions are detailed in \Secref{app:streaming_details}; bounded memory assembly and prompt adaptation are specified in \Secref{app:bank_assembly}.
The resulting latency and memory overhead are measured in \Secref{app:efficiency}.

\begin{table}[t]
\centering
\caption{\textbf{Quantitative comparison of interactive long video generation.} 
All models are evaluated using their official checkpoints. The best result in each column is shown in \textbf{bold}, and the second best is \underline{underlined}. 
}
\vspace{-3mm}
\resizebox{\linewidth}{!}{
\begin{tabular}{l *{8}{c}}
\toprule
\multirow{2}{*}{\textbf{Method}} &
\multicolumn{1}{c}{\multirow{2}{*}{\makecell{Quality\\Score $\uparrow$}}} &
\multicolumn{1}{c}{\multirow{2}{*}{\makecell{Dynamic\\Degree $\uparrow$}}} &
\multicolumn{6}{c}{Text Align $\uparrow$} \\
\cmidrule(lr){4-9}
& & & \multicolumn{1}{c}{0--10\,s} & \multicolumn{1}{c}{10--20\,s} &
\multicolumn{1}{c}{20--30\,s} & \multicolumn{1}{c}{30--40\,s} &
\multicolumn{1}{c}{40--50\,s} & \multicolumn{1}{c}{50--60\,s} \\
\midrule
Infinity-RoPE~\citep{yesiltepe2025infinity} & 79.99 & 36.33 & 69.58 & 66.20 & 65.38 & 64.51 & 63.81 & 59.98 \\
LongLive~\citep{yang2025longlive} & 79.09 & 28.30 & 71.88 & \textbf{67.97} & \textbf{67.41} & \textbf{66.36} & \textbf{65.91} & 62.29 \\
MemFlow~\citep{ji2025memflow} & 78.65 & 26.05 & \underline{71.95} & 66.74 & 65.79 & 64.95 & 64.03 & 60.29 \\
IAMFlow~\citep{iamflow} & 78.82 & 24.30 & \textbf{72.41} & 66.91 & 65.69 & 64.62 & 64.26 & 59.87 \\
Anchor Forcing~\citep{yang2026anchorforcing} & \textbf{82.32} & \underline{72.87} & 71.83 & \underline{67.83} & \underline{65.82} & 65.28 & 64.70 & \underline{62.41} \\
\textbf{\ours{}} & \underline{81.61} & \textbf{81.38} & 70.83 & 67.02 & 65.38 & \underline{66.16} & \underline{65.09} & \textbf{62.97} \\
\bottomrule
\end{tabular}
}
\label{tab:interactive}
\vspace{-5mm}
\end{table}

\subsection{Slot-Based Semantic KV Compression}
\label{sec:method_writing}

The first component converts completed-chunk history into entries addressable by meaning rather than position.
A naive compression stores low-dimensional token summaries, but diffusion attention directly consumes K/V features, and projected summaries cannot be recovered without loss.
\ours{} therefore separates memory \emph{address} and \emph{content}: slots provide compact semantic addresses, while stored content remains original transformer K/V.

For each chunk, clean key tokens are flattened over heads and projected into a slot-aligned latent space by adapter $\phi_l$, so that each slot assigns complete tokens rather than individual heads:
\begin{equation}
Z_l^c=\phi_l(\bar K_l^c),\qquad
(U_l^c,A_l^c)=\operatorname{SlotAttn}(Z_l^c;\bar U_l^{c-1}).
\end{equation}
Here $U_l^c$ denotes slot descriptors and $A_l^c$ their soft token assignments.
Slot Attention initializes from previous slot state $U_l^{c-1}$ when available, otherwise a learned initializer $U_l^0$, enabling consecutive chunks to share a slot vocabulary.

A slot defines a semantic memory unit but is read in the original K/V space, so each slot stores $R$ assigned tokens verbatim instead of descriptors.
Selected states are gathered as
\begin{equation}
\label{eq:slot_address_content}
(\hat K_l^{c,s},\hat V_l^{c,s})=
\big(\bar K_l^c[I_{l,s}^c],\bar V_l^c[I_{l,s}^c]\big),
\end{equation}
where $I_{l,s}^c$ contains the $R$ selected token indices; memory entries thus remain attention-compatible K/V groups rather than compressed embeddings.
The token-selection score, write-quality estimate, and candidate construction are given in \Secref{app:slot_details}.

\subsection{Persistent Entity Bank}
\label{sec:method_entitybank}

\begin{table}[t]
  \setlength{\tabcolsep}{3.5pt}
  \footnotesize
  \caption{
    \textbf{Quantitative comparisons under single-prompt 5-second setting.} 
    Among autoregressive models, the best result in each metric column is shown in \textbf{bold}, and the second-best is \underline{underlined}. 
  }
  \vspace{-3mm}
  \label{tab:short}
  \centering
\resizebox{\linewidth}{!}{
\begin{tabular}{lcccccc}
  \toprule
  \multirow{2}{*}{Model} & \multirow{2}{*}{\#Params} & \multirow{2}{*}{Resolution} & \multirow{2}{*}{Dynamic $\uparrow$} & \multicolumn{3}{c}{Evaluation scores $\uparrow$}\\
  \cmidrule(lr){5-7}
   &  &  &  & Total & Quality & Semantic \\
  \midrule
  \rowcolor{catgray}
  \multicolumn{7}{l}{\textit{Diffusion models}}\\
  LTX-Video~\citep{hacohen2024ltx}      & 1.9B & $768{\times}512$ & 75.00 & 82.33 & 85.20 & 70.83 \\
  Wan2.1~\citep{wan2025wan}                     & 1.3B & $832{\times}480$ & 73.61 & 84.27 & 85.24 & 80.43 \\
  \midrule
  \rowcolor{catgray}
  \multicolumn{7}{l}{\textit{Autoregressive models}}\\
  SkyReels-V2~\citep{chen2025skyreels}       & 1.3B & $960{\times}540$ & 48.61 & 81.72 & 82.93 & 76.90 \\
  CausVid~\citep{yin2025slow}             & 1.3B & $832{\times}480$ & 63.89 & 82.96 & 83.94 & 79.02 \\
  Pyramid Flow~\citep{jin2025pyramidal}      & 2B   & $640{\times}384$ & 68.06 & 81.56 & 83.89 & 72.24 \\
  Self-Forcing~\citep{huang2026self}         & 1.3B & $832{\times}480$ & 63.89 & 83.83 & 84.60 & 80.79 \\
  Infinity-RoPE~\citep{yesiltepe2025infinity}& 1.3B & $832{\times}480$ & 58.33 & 83.55 & 84.34 & 80.43 \\
  DeepForcing~\citep{yi2025deep}             & 1.3B & $832{\times}480$ & 63.89 & \underline{83.85} & \underline{84.61} & 80.84 \\
  MemFlow~\citep{ji2025memflow}              & 1.3B & $832{\times}480$ & 50.00 & 81.76 & 83.74 & 73.86 \\
  LongLive~\citep{yang2025longlive}          & 1.3B & $832{\times}480$ & 37.50 & 82.81 & 83.26 & \textbf{81.01} \\
  Anchor Forcing~\citep{yang2026anchorforcing}          & 1.3B & $832{\times}480$ & \textbf{73.61} & \textbf{83.99} & \textbf{84.84} & 80.57 \\
  \midrule
  \textbf{\ours{}}          & 1.3B & $832{\times}480$ & \underline{70.27} & 83.68 & 84.38 & \underline{80.85} \\
  \bottomrule
\end{tabular}
}
\vspace{-5mm}
\end{table}

Slot initialization ties a chunk's semantic decomposition to its predecessor, but slots are not identities: an entity may map to different slots after gaps, and slots may be reassigned after prompt changes.
Entity memory complements slot memory by maintaining, independently per semantic layer, a reusable identity bank $\mathcal E_l^c=\{e_j\}_{j=1}^{J}$ across the video.
Each entity $e_j$ stores an identity key $\kappa_j$ for slot comparison, a quality estimate, lifecycle state, and bounded transformer-compatible K/V snapshots.
Association determines slot-to-identity correspondence for the latest completed chunk.
Given slot descriptors $U_{l,s}^{c-1}$ from chunk $c-1$ and keys in $\mathcal E_l^{c-1}$, admissible pairs are scored by cosine similarity and assigned under a gate:
\begin{equation}
\operatorname{sim}_{s,j}^c=\cos\!\big(U_{l,s}^{c-1},\kappa_j^{c-1}\big).
\end{equation}
Hungarian matching solves the gated assignment: matched slots reinforce existing identities, while unmatched slots with sufficient write quality initialize tentative entities.
The matching gate and lifecycle transitions are specified in \Secref{app:entity_details}.

An identity requires visual content to condition generation.
Each entity stores bounded K/V history from representative slot-compressed tokens, so its retrieved snapshots remain original attention-compatible K/V rather than embeddings.
Retrieval occurs before current slot commitment, so chunk $c$ accesses history through chunk $c-1$:
\begin{equation}
\mathcal R_l^c=\operatorname{Retrieve}(U_l^{c-1},\mathcal E_l^{c-1}),\qquad
\mathcal B_l^c=\operatorname{Assemble}(\mathcal R_l^c,\mathcal H_l^{c-1},\mathcal W_l^{c-1}).
\end{equation}
Returned groups are inserted into attention by the assembly step in \Secref{app:bank_assembly}; current slots are then committed for later chunks.
Retrieval, snapshot retention, and persistent access to temporarily absent entities are detailed in \Secref{app:entity_details}: continuity for an entity that re-enters after a long absence comes from restoring its stored snapshots into the bank rather than from re-identifying the returning slot.
Retrieved snapshots retain the rotary phases under which they were written; \Secref{app:rope} examines this design choice.

\paragraph{Training.}
Training must expose memory to self-generated states, since the model writes inference-time entries during rollout.
Each iteration continues this rollout: the generator synthesizes chunks sequentially, writes each completed chunk's clean state into memory, and conditions future chunks on stored states.
The generator is adapted by distribution matching distillation (DMD), where a trainable fake-score network and frozen teacher score the same noisy sample, and their difference guides generator updates.

The objective combines this term with memory regularization:
\begin{equation}
\mathcal L_{\mathrm{gen}}=\mathcal L_{\mathrm{DMD}}+\lambda_{\mathrm{slot}}\mathcal L_{\mathrm{slot}}.
\end{equation}
These terms address slot drift, inaccurate content representation, and token selection unrelated to actual chunk usage.
Full implementation details are collected in \Secref{app:impl}, which specifies the streaming rollout (\Secref{app:training_rollout}), the two-stage schedule and frozen modules (\Secref{app:training_schedule}), the loss definitions (\Secref{app:training_details}), and the optimization recipe (\Secref{app:training}).

%% file: sec/4_exp.tex
\section{Experiments}
\label{sec:experiments}

\begin{figure*}[t]
\centering
\includegraphics[width=1.0\textwidth]{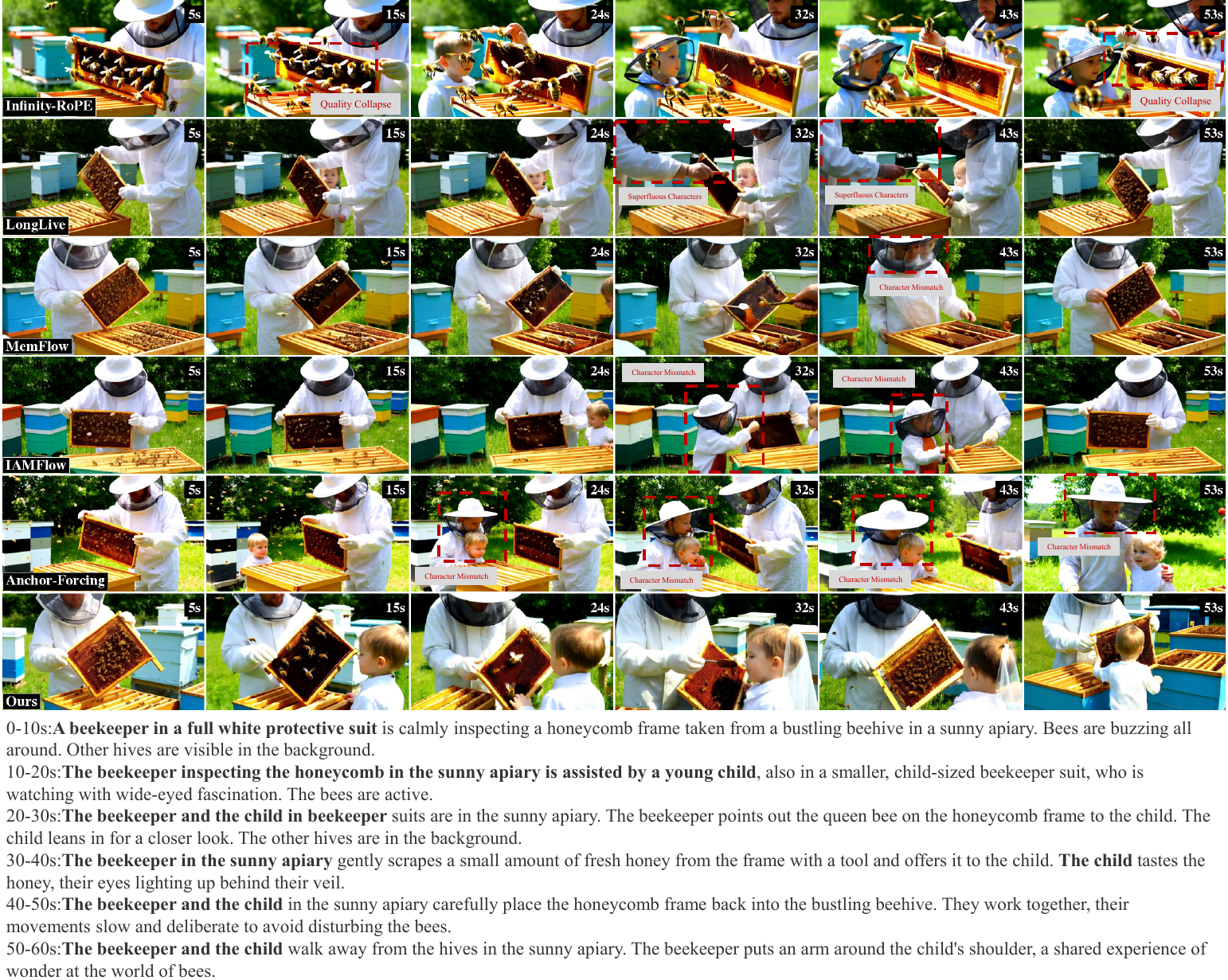}
\vspace{-5mm}
\caption{\textbf{Qualitative comparison of interactive long-form text-to-video generation.} 
}
\vspace{-5mm}
\label{fig:interactive_comparison_grid}
\end{figure*}

\subsection{Experiment Setups}
\noindent\textbf{Implementation Details.}
\label{sec:exp_implementation}
We instantiate \ours{} using the pretrained Wan2.1-T2V-1.3B backbone~\citep{wan2025wan}, 
which generates 5-second clips at 16 FPS and a resolution of 832 $\times$ 480. 
Following the optimization strategy of LongLive~\citep{yang2025longlive} and MemFlow~\citep{ji2025memflow}, 
the training process is divided into two phases. 
First, the base model is adapted into a few-step causal-attention generator via a Self-Forcing Distribution Matching Distillation pipeline~\citep{huang2026self} on the VidProM dataset~\citep{wang2024vidprom}, 
utilizing short-window attention and first-frame sink tokens to stabilize autoregressive rollouts. 
Second, we freeze the backbone weights and initialize the slot adapter and memory modules for streaming long-tuning. 
This phase utilizes 60-second multi-prompt sequences where each iteration continues the model's own rollout. 
Prompt switches are sampled uniformly between 5 and 55 seconds to train the module to perform both entity preservation and context suppression. 
The memory bank and sparse activation modules are optimized for 3,000 steps using AdamW (learning rates of $1.0 \times 10^{-5}$ for the actor and $2.0 \times 10^{-6}$ for the critic). 
Training is completed in approximately 12 hours on 8 NVIDIA H200 GPUs.

\noindent\textbf{Benchmarks and Datasets.}
\label{sec:exp_datasets}
To evaluate temporal robustness across varying horizons,
we establish benchmarks for three distinct scenarios:
short-video generation (5s),
long single-prompt generation (30s,120s,240s),
and interactive multi-prompt narratives (60s).
For 5s and long single-prompt evaluations,
we adopt standard protocols from Self-Forcing~\citep{huang2026self} and LongLive~\citep{yang2025longlive}.
The 60s interactive benchmark follows MemFlow~\citep{ji2025memflow},
comprising 100 narrative scripts.
Each script contains six consecutive 10s prompts,
requiring the model to maintain semantic consistency across five discrete transitions.

\noindent\textbf{Evaluation Metrics.}
\label{sec:exp_metrics}
We perform a multidimensional quantitative assessment using the VBench~\citep{huang2023vbench} and VBench-Long~\citep{huang2025vbench++} frameworks. 
Short-video performance is measured via Total, Quality, Semantic, and Dynamic scores. 
For the 30-second horizon, we extend this to include Imaging Quality and Temporal Style to capture long-range perceptual stability. 
In the 60-second interactive setting, 
we report the aggregate Quality score to reflect overall visual fidelity. 
To specifically analyze the impact of prompt transitions, 
we compute segment-wise text align use Qwen3-VL-Embedding-8B~\citep{qwen3vl} for each 10-second interval, 
providing a granular view of how well semantic identities are preserved as the narrative evolves.

\begin{table}[t]
\centering
\caption{\textbf{Quantitative comparisons under a 30-second single-prompt setting.} We evaluate extended-horizon generation capabilities against recent autoregressive streaming models ($\uparrow$ higher is better). The best result in each metric column is shown in \textbf{bold}, and the second-best is \underline{underlined}. 
}
\vspace{-3mm}
\label{tab:long}
\setlength{\tabcolsep}{6pt}
\renewcommand{\arraystretch}{1.15}
\resizebox{1.0\linewidth}{!}{
\begin{tabular}{lcccccc}
\toprule
\textbf{Model} & \makecell{Total\\Score $\uparrow$} & \makecell{Quality\\Score $\uparrow$} & \makecell{Semantic\\Score $\uparrow$} & \makecell{Dynamic\\Score $\uparrow$} & \makecell{Imaging\\Quality $\uparrow$} & \makecell{Temporal\\Style $\uparrow$}\\
\midrule
Self-Forcing~\citep{huang2026self} & 82.21 & 83.38 & 77.54  & 51.85 & 68.17 & 23.31\\
Infinity-RoPE~\citep{yesiltepe2025infinity} & 83.16 & 84.17 & 79.13  & 57.04 & 67.94 & 23.79\\
DeepForcing~\citep{yi2025deep}  & 82.26 & 83.20 & 78.50 & 52.86 & 67.65 & 23.51\\
MemFlow~\citep{ji2025memflow}     & 82.95 & 83.86 & 79.31 & 60.46 & 67.68 & 23.89\\
LongLive~\citep{yang2025longlive}     & 82.77 & 83.31 & \textbf{80.64} & 40.19 & 68.96 & 23.97\\
Anchor Forcing~\citep{yang2026anchorforcing}     & \underline{83.25} & \underline{84.40} & 78.67 & \underline{79.17} & \textbf{69.37} & \textbf{24.17}\\
\textbf{\ours{}}     & \textbf{83.79} & \textbf{84.86} & \underline{79.49}  & \textbf{79.30} & \underline{69.11} & \underline{24.09}\\
\bottomrule
\end{tabular}}
\vspace{-5mm}
\end{table}

\subsection{Main Experiments}
\label{sec:exp_main}

\noindent\textbf{Interactive Multi-Prompt Generation.}
Prompt switches are the sharpest test of what a memory keeps, because the instruction changes while the entities it describes do not.
As shown in Table~\ref{tab:interactive}, \ours{} attains the best Dynamic Degree by a wide margin (81.38 against 72.87 for the next best method) while ranking second on aggregate Quality (81.61 against 82.32 for Anchor Forcing); LongLive and MemFlow stay below 80 on Quality and below 30 on Dynamic.
The segmented text alignment shows that this margin is not an artifact of one segment: \ours{} is not the best method early on, but it is the most stable one across the sequence, losing 7.86 points from the first segment to the last, against 9.4--12.5 points for the baselines, and it ends on the highest score of any method.
A memory that carries entities across switches should decay more slowly than one that rebuilds the scene from the recent window at each switch, and the separation appears exactly where such a delay accumulates: at the end of the sequence.
The qualitative grid in \Figref{fig:interactive_comparison_grid} shows the same pattern on a six-prompt apiary script, where the baselines lose the content or the cast while \ours{} keeps the beekeeper, the child, and the frame consistent throughout.
Additional scripts, covering a cast change and two stress cases with annotated failure modes, are shown in \Secref{app:qualitative}.

\noindent\textbf{Long-Video Generation.}
At 30 seconds the rollout is long enough for accumulated drift to dominate, and this is the setting where slot memory separates from the temporal memory it replaces.
Table~\ref{tab:long} places \ours{} first on Total (83.79), Quality (84.86), and Dynamic (79.30) and second on the remaining three dimensions, with no other method leading on more than two.
Dynamic Degree is the diagnostic column: MemFlow and LongLive fall to 60.46 and 40.19, while \ours{} holds 79.30.
A temporal memory preserves consistency by re-using recent content, which ties how much a scene can move to how much of it is replayed; a memory that stores entity appearance instead of scene content leaves the scene free to move.
The nearest competitor, Anchor Forcing, matches \ours{} to within 0.6 points on Total and Quality, so the methods are separated not by perceptual quality but by what each memory keeps: ours stays close to LongLive on Semantic, while Anchor Forcing drops to 78.67.
The same separation persists at 120\,s and 240\,s, where appearance metrics saturate while the margin in Dynamic Degree widens; see \Secref{app:longhorizon} and Table~\ref{tab:main}.

\noindent\textbf{Short-Video Generation.}
A 5-second clip is too short for the bank to be exercised, so the question here is not whether memory helps but whether attaching it costs anything.
It does not (Table~\ref{tab:short}): \ours{} lands within 0.6 and 0.9 points of the Wan2.1-T2V-1.3B backbone on Total and Quality, and among autoregressive models it is second on both Dynamic and Semantic.
The Dynamic column again carries the mechanism: MemFlow and LongLive lose motion at this horizon, whereas \ours{} is higher than every autoregressive baseline except Anchor Forcing.
This is what the design predicts, since a bank that only becomes useful once the early frames leave the recent window should be close to inert when no such history exists yet.

\noindent\textbf{Summary of Scaling Performance.}
The three horizons give one consistent picture: at 5 seconds \ours{} behaves like the backbone it extends, at 30 seconds it wins the dimensions that depend on memory, and at 60 seconds it is the most stable method across five prompt switches.
The margin over the baselines therefore grows with horizon and concentrates in motion, appearance, and late-segment prompt adherence rather than in short-clip fidelity, which is what the persistent memory protects and what the baselines lose first.
Extended quantitative results, including the long-horizon and efficiency measurements, are collected in \Secref{app:quantitative}.

\begin{table*}[t]
  \centering
  \setlength{\tabcolsep}{3pt}
  \caption{
  \textbf{Core mechanism ablation.}
  Tail alignment averages 30--60s; text drift is the 0--10s to 50--60s drop.
  FG-ARI and mBO use ReferDINO~\citep{ReferDINO} pseudo-masks from prompt-extracted object text, averaged over the top three layers.
  Bold and underline mark the best and second-best results.
  }
  \vspace{-3mm}
  \label{tab:core_ablation}
  \begin{tabular}{lcccccc}
    \toprule
    Method
      & \makecell{Text\\avg. $\uparrow$}
      & \makecell{Tail\\align. $\uparrow$}
      & \makecell{Text\\Drift$\downarrow$}
      & \makecell{FG-ARI $\uparrow$}
      & \makecell{mBO $\uparrow$}
      & \makecell{VBench\\quality $\uparrow$} \\
    \midrule
    Baseline w/o Slot Compression
      & 65.63 & 63.09 & 11.66 & -- & -- & 78.65 \\
    + Slot (no entity bank)
      & 66.15 & 63.71 & 11.17 & 8.01 & 9.42 & 78.89 \\
    + Entity Bank (w/o $\mathcal{L}_{\mathrm{dist}}$)
      & \textbf{66.63} & \textbf{65.12} & \textbf{7.84} & \underline{10.42} & \textbf{15.64} & \underline{81.50} \\
    Top-$k$ KV Memory
      & 65.66 & 63.58 & 10.23 & -- & -- & 80.59 \\
    Ours (full)
      & \underline{66.24} & \underline{64.74} & \underline{7.86} & \textbf{11.77} & \underline{15.53} & \textbf{81.61} \\
    \bottomrule
  \end{tabular}%
\end{table*}
\begin{figure}[t]
    \centering
    \includegraphics[width=1\linewidth]{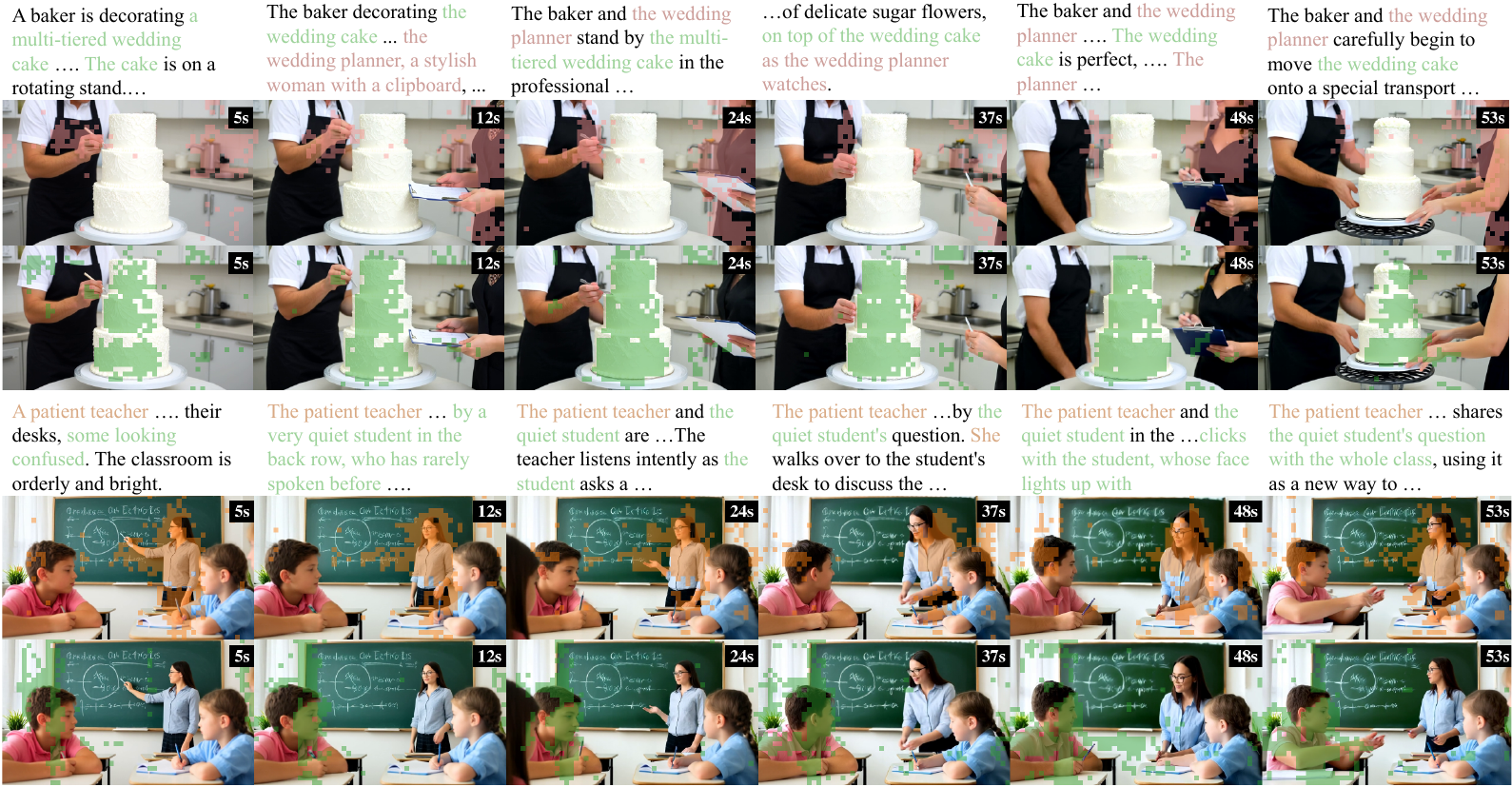}
    \vspace{-5mm}
    \caption{\textbf{Slot-wise representative tokens.}
Colored patches mark retained tokens for different slots across two sequences; slots track distinct objects and scene regions over time.}
\vspace{-5mm}
    \label{fig:slot_vis}
\end{figure}

\paragraph{User study.} We conducted a human evaluation of long-form video quality across identity preservation, action completion, scene consistency, temporal stability, and object interaction. Twenty users compared 300 video groups and preferred our method in 81 cases (27\%), versus Anchor-Forcing in 62 (20.7\%), IAMFlow in 37 (12.3\%), Infinity-RoPE in 43 (14.3\%), LongLive in 52 (17.3\%), and MemFlow in 25 (8.3\%); \ours{} was the most frequently preferred method in this comparison. The protocol, evaluation criteria, and quality-control rules are given in \Secref{app:human_evaluation}.
Latency, peak memory, and throughput under the interactive protocol are reported in \Secref{app:efficiency} (Table~\ref{tab:efficiency}).

\newpage
\subsection{Ablation Studies}

\begin{wrapfigure}[15]{r}{0.5\linewidth}
    \centering
    \vspace{-5mm}
    \includegraphics[width=\linewidth]{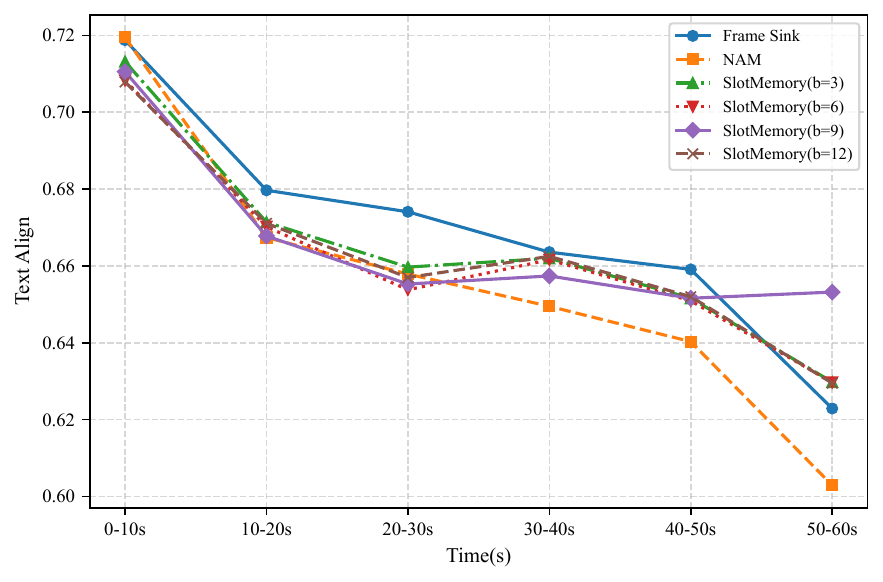}
    \vspace{-5mm}
    \caption{\textbf{Segment-wise text alignment across memory designs and bank sizes.}}
    \label{fig:clipscore_banksize}
\end{wrapfigure}
We ablate in the 60-second interactive setting of Table~\ref{tab:interactive}, varying one axis at a time---what the memory is made of, what supervises the slots, and how much memory is kept---while holding the backbone, local window, chunk protocol, prompt-switch protocol, and retained-token budget fixed.

\begin{table*}[t]
  \centering
  \small
  \setlength{\tabcolsep}{4pt}
  \caption{
  \textbf{Loss ablation under the full \ours{} mechanism (slot compression + persistent entity bank).}
  Tail text alignment is averaged over 30--60s.
  Best and second-best results are highlighted in \textbf{bold} and \underline{underlined}, respectively.
  }
  \vspace{-3mm}
  \label{tab:loss_ablation}
  \begin{tabular}{lccccc}
    \toprule
    Slot regularization
      & \makecell{Subject\\consistency $\uparrow$}
      & \makecell{Background\\consistency $\uparrow$}
      & \makecell{Dynamic\\degree $\uparrow$}
      & \makecell{VBench\\quality $\uparrow$}
      & \makecell{Tail text\\align. $\uparrow$} \\
    \midrule
    $\mathcal{L}_{\mathrm{con}}+\mathcal{L}_{\mathrm{rec}}$ (w/o $\mathcal{L}_{\mathrm{dist}}$)
      & 93.07 & 92.76 & \textbf{92.96} & \underline{81.50} & \textbf{65.12} \\
    $\mathcal{L}_{\mathrm{con}}+\mathcal{L}_{\mathrm{dist}}$ (w/o $\mathcal{L}_{\mathrm{rec}}$)
      & \textbf{95.36} & \textbf{93.90} & 59.08 & 79.99 & 63.24 \\
    Complete ($\mathcal{L}_{\mathrm{con}}+\mathcal{L}_{\mathrm{rec}}+\mathcal{L}_{\mathrm{dist}}$)
      & \underline{95.00} & \underline{93.86} & \underline{81.38} & \textbf{81.61} & \underline{64.74} \\
    \bottomrule
  \end{tabular}
  \vspace{-2mm}
\end{table*}
\noindent\textbf{Impact of Slot Compression and Entity Persistence.}
Table~\ref{tab:core_ablation} separates semantic addressing from entity persistence.
Slot compression groups retained tokens by entity but barely changes aggregate scores without cross-chunk identity matching.
Adding the entity bank reduces text drift from 11.17 to 7.84 and improves tail alignment, mBO, and FG-ARI, while average alignment changes little.
We analyze the identity key's positional representation in \Secref{app:rope}.

\noindent\textbf{Structure versus Budget.}
The Top-$k$ row separates the retained tokens from the addresses they are stored under.
Untyped tokens of the same budget recover most of the compression loss in overall quality, so the budget is not the bottleneck; what they cannot express is which token belongs to which entity, and the variant consequently behaves like slot-only memory at the end of the sequence.
Slots are what make the tokens addressable: they are the unit the entity bank associates, the unit a prompt switch retrieves by, and the unit bounded assembly selects from. \Figref{fig:slot_vis} shows that grouping directly, and the FG-ARI and mBO columns of Table~\ref{tab:core_ablation} quantify its coherence over time.

\noindent\textbf{What the Two Slot Losses Supervise.}
The two regularizers in Table~\ref{tab:loss_ablation} are not interchangeable.
Removing reconstruction leaves the highest subject and background consistency but takes the motion with it---slots that are no longer grounded in the tokens the backbone renders collapse Dynamic Degree from 93 to 59---whereas removing distillation keeps motion and the best tail alignment but gives up consistency and slot--object alignment.
Neither term dominates, and only the complete objective reaches the best overall VBench quality without collapsing on either axis.

\noindent\textbf{Capacity and Memory Design.}
Bank capacities differ mainly in the final segment: nine blocks improve alignment and motion, while the six-block default yields the highest overall quality.
In \Figref{fig:clipscore_banksize}, frame-sink caching and NAM match slot memory early but lose roughly twice as much alignment by the end, when entity recall matters most.

%% file: sec/5_conclusion.tex
\section{Conclusion}
We introduce \ours{}, an object-centric KV memory that uses semantic slots to retrieve original transformer KV tokens and maintain entity continuity across streaming chunks and prompt switches. On 60-second interactive narratives, it achieves the highest Dynamic Degree and final-segment text alignment among evaluated methods while ranking second in aggregate Quality. At 30 seconds, it leads in Total, Quality, and Dynamic scores, with 31.2\% higher Dynamic Degree than MemFlow. Ablations further show that persistent entity tracking, beyond slot compression alone, reduces late-stage text drift and improves object-level coherence. These results indicate that addressable entity memory is what preserves motion and identity at long horizons.

%% file: sec/6_appendix.tex
\appendix

\setcounter{table}{0}
\setcounter{figure}{0}
\renewcommand{\thetable}{\thesection.\arabic{table}}
\renewcommand{\thefigure}{\thesection.\arabic{figure}}

\section{Extended Related Work}
\label{app:related}

This section supplies the broader literature and mechanism-level details behind the compressed summary in
\Secref{sec:related}. It expands the citation base of the two lines that the main text only sketches,
and states where each line stops short of the memory primitive used by \ours{}.

\subsection{Streaming Video Generation and Memory Mechanisms}
\label{app:related_video}

\noindent\textbf{From clip synthesis to streaming rollouts.}
Early text-to-video diffusion models generate isolated clips~\citep{ho2022videodiffusion,singer2022makeavideo,ho2022imagenvideo,wu2022tuneavideo,guo2023animatediff,blattmann2023svd,wang2023modelscope,bar2024lumiere,hacohen2024ltx,wan2025wan,jin2025pyramidal,chen2025skyreels}.
Because each rollout remains bounded, long-video generation has converged on autoregressive streaming frameworks that synthesize chunks under a fixed budget and carry state across chunks through KV caches.
These systems differ mainly in how they stabilize the rollout: FIFO-Diffusion generates infinite video without training by a queue-based pipeline~\citep{kim2024fifo};
StreamingT2V adds appearance and motion conditioning for extendable narratives~\citep{henschel2025streamingt2v};
CausVid distills a bidirectional model into a few-step causal generator~\citep{yin2025slow};
Self-Forcing trains on self-generated rollouts to close the train--test gap~\citep{huang2026self};
Infinity-RoPE controls positional phases for unbounded, action-controllable rollouts~\citep{yesiltepe2025infinity};
Deep Forcing combines deep sink tokens with participative compression in a training-free manner~\citep{yi2025deep};
\citet{gu2025far} cast the problem as next-frame prediction;
FramePack packs frame context and suppresses drift~\citep{zhang2025framepack};
TTT-Video updates a test-time memory for one-minute generation~\citep{dalal2025tttvideo};
and Magi-1~\citep{teng2025magi1} and Inflvg~\citep{fang2025inflvg} scale or reinforce autoregressive rollouts.

\noindent\textbf{Memory organization.}
Where these frameworks keep history, they do so through explicit banks.
LongLive maintains an interactive KV cache with sink protection~\citep{yang2025longlive};
Context-as-memory retrieves scene-consistent context for interactive generation~\citep{yu2025context};
Mixture-of-Contexts composes multiple contexts for long video~\citep{cai2025mixture};
MemFlow adapts the memory flow to the narrative~\citep{ji2025memflow};
and Anchor Forcing combines anchor memory with a tri-region RoPE~\citep{yang2026anchorforcing}.
The memory mechanisms summarized in \Secref{sec:related} differ in what they address and what they store.
LongLive-RAG retrieves self-generated latents with a learned retrieval embedding and a windowed temporal-delta loss~\citep{hu2026longliverag};
OmniMem selects a sparse subset of the full historical KV cache and shares the selection across heads~\citep{zhao2026omnimem};
FadeMem consolidates KV blocks into a distance-aware hierarchy that keeps near history dense and far history sparse~\citep{lu2026fademem};
MemRoPE compresses past keys into long- and short-term exponential-moving-average tokens and re-indexes them online with RoPE~\citep{kim2026memrope};
SWIFT injects prompt-adaptive segment anchors into a semantic cache~\citep{tan2026swift};
Echo-Forcing maintains scene-level anchors and recall frames with difference-aware decay~\citep{wu2026echo};
EM-Vid indexes entity-relevant latent patches and attends only to them, without training~\citep{vandersanden2026emvid};
and Memento trains subject reconstruction together with a dual-query memory~\citep{wei2026memento}.
What distinguishes \ours{} from this spectrum is not the budget but the addressing content: these designs address frames, chunks, scenes, or externally stored summaries, and the entity-centric designs listed above index entities outside the attention stack; \ours{} instead assigns entity-level addresses inside the diffusion transformer and keeps the stored values as original KV tokens.

\subsection{Object-Centric Representation Learning}
\label{app:related_slots}

Foundational object-centric models decompose a scene into slots~\citep{burgess2019monet,greff2019iodine,engelcke2019genesis,locatello2020object}.
The subsequent literature extends them along four directions that are only summarized in \Secref{sec:related}.
First, temporal and real-world video: conditional video slot attention~\citep{kipf2022conditional},
SAVi++ for end-to-end learning from real videos~\citep{elsayed2022savipp},
SlotFormer for visual dynamics simulation~\citep{wu2023slotformer},
and DINOSAUR-style models that scale object-centric learning to natural images and their successors~\citep{seitzer2023dinosaur,didolkar2025transfer},
with temporally consistent slot contrast~\citep{manasyan2025temporally} and grounded or noise-regularized variants~\citep{kori2024grounded}.
Second, geometric structure: slot-conditioned diffusion decoders~\citep{jabri2024dorsal},
object-centric voxelization for dynamic scenes~\citep{zhao2024dynavol},
and movable-object inference for 3D scene representation~\citep{chen2024morf}.
Third, identifiability and disentanglement: probabilistic slot attention~\citep{kori2024probslot}
and explicitly disentangled slot representations~\citep{majellaro2024disentangled},
together with recent work that separates identity from pose or static from dynamic factors~\citep{wei2026staitus,zhao2026xssc}
and object-centric world models~\citep{zhang2025objectsworldmodels}.
Fourth, generative and language conditioning: language-controllable slot inference~\citep{didolkar2025ctrlo}
and diffusion-coupled slots with register tokens and contrastive alignment~\citep{nguyen2026coda}.
Across all four directions, the slot is the representation that is reconstructed, predicted from, or decoded;
memory is therefore bounded by the fidelity of that representation.
\ours{} inverts this arrangement by keeping the content in the original KV space and letting slots act only as routing keys,
so the same semantic decomposition can be queried by attention without a low-dimensional reconstruction bottleneck.

\section{Method Specification}
\label{app:method}

This appendix specifies the exact instantiations, ordering, and edge-case behavior that
\Secref{sec:method} abstracts over. Equations already given there are referenced rather than repeated,
and all values are the defaults used in the reported experiments.

\subsection{Notation and Architecture Placement}
\label{app:notation}

\noindent\textbf{Units.}
A \emph{token} is one patchified latent position.
With patch size $(1,2,2)$ on a $60{\times}104$ latent grid, one latent frame contains $1{,}560$ tokens,
so a generated chunk of $T_c{=}3$ latent frames contains $N{=}4{,}680$ tokens.
A \emph{slot} is one of $S{=}7$ semantic groups produced per written chunk,
described by a $d_s{=}64$-dimensional address and $R{=}64$ selected original K/V tokens.
An \emph{entity} is a persistent identity with a $d_s$-dimensional key, a scalar quality estimate,
a lifecycle state, and up to $M{=}4$ stored K/V snapshots.
The Wan2.1-T2V-1.3B backbone~\citep{wan2025wan} uses $L{=}30$ layers with $H{=}12$ attention heads and $D{=}128$ channels per head
($H{\cdot}D{=}1{,}536$); the batch size is $B{=}1$ in all reported runs.

\noindent\textbf{States.}
\emph{Local cache} $\mathcal C_l^c$ holds a fixed sink prefix (3 frames)~\citep{yang2025longlive} and a rolling window (12 frames).
\emph{Bank} $\mathcal B_l^c$ is the attention-visible memory: on semantic layers it holds at most
$N_{\mathrm{budget}}{=}6$ memory blocks of $7$ groups$\times64$ tokens, i.e.\ $42$ groups
($2{,}688$ K/V tokens); on raw layers it holds the most recent $3{\times}1{,}560{=}4{,}680$ tokens.
\emph{Ordinary history} $\mathcal H_l^c$ is a notational abstraction for the non-retrieved part of the bank;
the implementation stores it inside the same K/V tensors rather than in a separate object.
A \emph{candidate} is the set of $S$ groups with their write scores produced by compressing one chunk,
and an \emph{entity snapshot} is one stored K/V group of $64$ tokens belonging to an entity.

\noindent\textbf{Layer placement.}
Semantic memory is active on layers $\ell\in[L/3,\,2L/3]=[10,20]$, i.e.\ $11$ of the $30$ layers
(the bound is inclusive in the implementation).
These layers run slot compression and entity tracking independently;
all remaining layers keep only raw KV windows and maintain neither slots nor entities.
Every layer owns its own cache, bank, slot state, and (where applicable) entity table.
Table~\ref{tab:appendix_shapes} summarizes the resulting tensor shapes.

\subsection{Streaming State and Causal Update Order}
\label{app:streaming_details}

Memory timing is controlled by two events.
A \emph{read event} occurs only on the \emph{first} denoising step of a chunk;
it consumes any pending clean K/V and rebuilds the bank before self-attention reads it.
A \emph{write event} occurs only for chunks whose starting frame index is a multiple of the write period;
with the default period of three chunks ($9$ latent frames),
one chunk in three stages its clean K/V as pending memory.
A chunk therefore never attends to memory it produced itself:
states written by chunk $c$ become visible at the first denoising step of chunk $c+1$,
and chunks outside the write schedule contribute to the local cache but not to the bank.

The implemented per-chunk order is:
\begin{enumerate}
\item Set the active prompt. At a switch point, reset the text-conditioning cache, recompute it with the new prompt,
and re-encode the most recent local context under the new prompt (up to $12$ frames at inference and up to $21$ during training);
the bank, slot state, and entity bank are left untouched.
\item At the first denoising step, consume the pending clean K/V from the previous write event.
On semantic layers, compress it, retrieve entity snapshots, assemble the bounded bank, and commit the new slots to the entity bank.
On raw layers, append the pending K/V to the raw window and truncate it to $4{,}680$ tokens.
\item Denoise the chunk while self-attention reads $[\text{sink};\text{bank};\text{local}]$.
Steps after the first reuse the same bank and do not touch memory.
\item Re-encode the generated chunk with clean context to obtain $(\bar K_l^c,\bar V_l^c)$,
and write these clean states into the local cache.
\item If chunk $c$ belongs to the write schedule, stage $(\bar K_l^c,\bar V_l^c)$ as pending memory;
otherwise the bank is unchanged.
\end{enumerate}

\begin{table}[t]
\centering
\caption{\textbf{Tensor shapes in \ours{}.} Values are the defaults of the reported runs.}
\label{tab:appendix_shapes}
\setlength{\tabcolsep}{6pt}
\renewcommand{\arraystretch}{1.1}
\begin{tabular}{lll}
\toprule
\textbf{Tensor} & \textbf{Shape} & \textbf{Value} \\
\midrule
Clean K/V of one chunk & $B\times N\times H\times D$ & $N{=}4{,}680$, $H{\cdot}D{=}1{,}536$ \\
Slot latent $Z_l^c$ & $B\times N\times d_s$ & $d_s{=}64$ \\
Slot descriptors $U_l^c$ & $B\times S\times d_s$ & $S{=}7$ \\
Assignments $A_l^c$ & $B\times S\times N$ & --- \\
Selected slot K/V & $B\times S\times R\times H\times D$ & $R{=}64$ \\
Bank K/V $\mathcal B_l^c$ & $B\times N_{\mathrm{bank}}\times H\times D$ & $N_{\mathrm{bank}}\le 2{,}688$ \\
Entity keys & $B\times E\times d_s$ & $E{=}32$ \\
Entity snapshots & $B\times E\times M\times R\times H\times D$ & $M{=}4$ \\
Write scores & $B\times S$ & --- \\
\bottomrule
\end{tabular}
\end{table}

\subsection{Slot Construction and K/V Token Selection}
\label{app:slot_details}

Compression operates on the clean K/V of a written chunk.
Heads are flattened so that a complete token, rather than an individual head, is assigned to a slot.
The input adapter $\phi_l$ is a two-layer MLP, $1{,}536\!\to\!4{,}096\!\to\!64$
(LayerNorm and ReLU after the first layer), and Slot Attention~\citep{locatello2020object} uses $S{=}7$ slots,
$2$ iterations, a GRU update, and $\varepsilon{=}10^{-6}$.
Slots are initialized from the previous chunk's clean slot descriptors (detached) when available~\citep{manasyan2025temporally},
otherwise from a learned initializer.
The output adapter $\psi_l$ is $64\!\to\!1{,}024\!\to\!1{,}024\!\to\!1{,}024\!\to\!1{,}536$ with ReLU activations.

For slot $s$ and token $t$, the implemented selection score is
$\rho_{s,t}=1.5\cos(\psi_l(u_s),\bar k_t)+0.3\log(A_{s,t}+10^{-6})$,
i.e.\ the reconstruction term is weighted $1.5$ and the assignment-log prior $0.3$.
Each slot keeps its top $R{=}64$ tokens by $\rho$, giving $448$ selected tokens per chunk;
the selected entries are gathered from the original K/V tensors.
Selection is hard: the top-$R$ indices are non-differentiable, while gradients still reach the selected K/V values;
the reconstruction loss uses a straight-through hard assignment whose forward pass takes the top-1 slot but whose backward pass flows through the soft weights.
No explicit tie-breaking rule is imposed, so ties follow the implementation's top-$k$ ordering.

Each slot also receives a write quality that combines coverage and reconstruction fidelity:
the coverage term is the slot's share of total assignment mass,
and the reconstruction term is the assignment-weighted cosine similarity between the decoded prototype and the clean keys,
mapped to $[0,1]$.
The two terms are combined as $w=\mathrm{clamp}(\text{coverage}+0.5\cdot\text{reconstruction},0,1)$
and used both for later retention and for entity admission.

\subsection{Entity Association and Lifecycle}
\label{app:entity_details}

Entity tracking runs independently per semantic layer.
Each layer maintains at most $E_{\max}{=}32$ entities, and each entity stores at most $M{=}4$ K/V snapshots.
Association compares the previous chunk's slot descriptors with entity keys by cosine similarity;
a small index prior ($0.03$) is added only for ACTIVE entities seen in the immediately preceding step.
Candidate pairs are filtered by a state-dependent gate and solved by Hungarian matching with a dummy column that absorbs unmatched slots.
By default, DORMANT entities are excluded from association,
so a returning entity is not re-identified; its stored snapshots remain retrievable instead.

\begin{table}[t]
\centering
\caption{\textbf{Entity-bank defaults.} Thresholds are cosine similarities unless noted.}
\label{tab:appendix_entity}
\setlength{\tabcolsep}{6pt}
\renewcommand{\arraystretch}{1.1}
\resizebox{\linewidth}{!}{%
\begin{tabular}{ll}
\toprule
\textbf{Parameter} & \textbf{Value} \\
\midrule
Entity capacity $E_{\max}$ / snapshots per entity $M$ & $32$ / $4$ \\
Association gate (ACTIVE, TENTATIVE) & $0.70$ \\
Association gate for DORMANT (re-ID disabled by default) & $0.78$ \\
TENTATIVE $\rightarrow$ ACTIVE & $2$ matched chunks \\
TENTATIVE removal & $>2$ consecutive unmatched chunks \\
ACTIVE $\rightarrow$ DORMANT & $2$ consecutive unmatched chunks \\
Identity-key EMA base rate / minimum similarity & $0.10$ / $0.70$ \\
Snapshot write: write quality / match similarity & $\ge 0.40$ / $\ge 0.50$ \\
Admission of an unmatched slot to TENTATIVE & write quality $\ge 0.40$ \\
\bottomrule
\end{tabular}}
\end{table}

Lifecycle decisions are per layer.
A new TENTATIVE entity is born when an unmatched slot has sufficient write quality,
unless it is a near-duplicate of an already matched entity (similarity $>0.97$ with low quality).
A TENTATIVE entity becomes ACTIVE after two matched chunks and is cleared after more than two consecutive unmatched chunks.
An ACTIVE entity becomes DORMANT after two consecutive unmatched chunks.
Only DORMANT entities and low-quality TENTATIVE entities are evictable;
among them, the lowest retention score is removed first, where retention combines quality, match count, and dormancy age.
ACTIVE entities are never evicted, and unmatched slots are discarded rather than replacing a confirmed identity when the table is full.

Matched entities update their identity key by an EMA whose rate is proportional to write quality and match similarity,
and only when that similarity exceeds $0.70$;
quality itself follows a $0.9/0.1$ EMA.
A snapshot is written when both write quality and match similarity pass the thresholds in Table~\ref{tab:appendix_entity};
when all $M$ snapshot slots are occupied, the snapshot with the lowest quality-plus-age score is replaced.
Snapshots store original K/V groups, so retrieval returns transformer-compatible content rather than embeddings.

\subsection{Retrieval, Memory Assembly, and Prompt Switching}
\label{app:bank_assembly}

Retrieval is performed only in no-gradient forwards and, in the implementation, before the current chunk's slots are committed;
chunk $c$ therefore queries the entity state that was available at chunk $c-1$.
Each candidate entity is scored by its best slot similarity plus small bonuses for quality, match count, and a current-chunk match;
snapshot scores add snapshot quality, a recency bonus that decays linearly over $50$ chunks, and a match bonus.
Selection proceeds in three quota rounds---matched entities, persistent non-matched entities (ACTIVE or DORMANT), and recent high-quality snapshots---
followed by a global backfill up to the history budget. At most one snapshot per entity is returned.
Because DORMANT entities are not matched but are eligible in the persistent round,
their snapshots can still be restored after the entity has left the scene.

The bank is assembled under the budget $N_{\mathrm{budget}}$ of \Secref{app:notation}.
When retained history plus the new candidates already fits, they are concatenated directly.
Otherwise, the new chunk contributes $7$ groups, leaving $42-7=35$ groups for history.
Retrieved entity snapshots take priority; the remaining budget is filled by history groups ranked by their stored write quality,
falling back to the text--visual score ($0.7$ text, $0.3$ visual) when no write score exists.
Retrieved and history groups are deduplicated by the cosine similarity of their mean-pooled keys:
a history group with similarity above $0.98$ to a retrieved snapshot is dropped and replaced by the next-ranked group.
The final bank is the concatenation of retrieved snapshots, retained history, and the new groups, in that order,
and its entries remain original K/V tokens.

At a prompt switch, the text-conditioning cache is recomputed and the recent local window is re-encoded under the new prompt;
the bank, slot state, and entity bank are preserved.
Thus the switch changes what the scene should become, while entity memory preserves what established entities should continue to look like.

\noindent\textbf{Positional handling.}
Clean keys are stored after rotary position embedding has been applied at their source positions,
and bank keys are concatenated with the sink and local segments without re-indexing.
Retrieved snapshots therefore keep the RoPE phases of the chunks in which they were written;
the implementation does not remap them to the current chunk's positions.

\noindent\textbf{Worked example.}
Consider an entity that is matched at chunks $1$--$3$ and then leaves the scene.
After two consecutive unmatched chunks it becomes DORMANT, which removes it from ordinary association but keeps its snapshots.
During the absence, its snapshots can still be returned by the persistent quota if their visual relevance is high.
When the entity reappears, the slot is again unmatched because DORMANT entities are not re-identified by default;
it either initializes a new TENTATIVE entity (if its write quality passes the admission threshold) or is discarded.
Visual continuity therefore comes from restoring the dormant snapshots into the attention bank,
not from re-associating the returning slot with the old identity---a behavior that also defines a limitation of the current implementation.

\subsection{Algorithm}
\label{app:algorithm}

Algorithm~\ref{alg:streaming_memory} summarizes the per-chunk procedure whose states and update order are specified in \Secref{app:streaming_details}.

\begin{algorithm}[t]
  \caption{Causal Streaming Generation with Hybrid Persistent Memory}
  \label{alg:streaming_memory}
  \begin{algorithmic}[1]
    \Require prompts $\{p^{(m)}\}_{m=1}^{M}$, switch points $\mathcal{S}$, causal generator $G$
    \Ensure video latents $V^1,\dots,V^{C}$
    \State initialize all local/text caches, the entity bank, and the memory bank to empty
    \For{$c=1$ \textbf{to} $C$}
      \State set the active prompt $p_c$ for chunk $c$
      \If{$c=1$}
        \State $\mathcal{C}_{l}^{\mathrm{text},1}\leftarrow\mathrm{EncodeText}(p^{(1)})$ for all $l$
      \ElsIf{chunk $c$ begins at a switch point $s_m$}
        \State $\mathcal{C}_{l}^{\mathrm{text},c}\leftarrow\mathrm{EncodeText}(p^{(m+1)})$ for all $l$
        \State recompute the recent local context under $p^{(m+1)}$ \Comment{the entity bank and persistent snapshots are retained}
      \Else
        \State $\mathcal{C}_{l}^{\mathrm{text},c}\leftarrow\mathcal{C}_{l}^{\mathrm{text},c-1}$ for all $l$
      \EndIf
      \If{$c>1$}
        \ForAll{semantic layers $l$}
          \State $(U_l^{c-1},\mathcal{W}_l^{c-1})\leftarrow\mathrm{Compress}\!\big(\bar{K}_l^{c-1},\bar{V}_l^{c-1},\bar{U}_l^{c-2}\big)$ \Comment{$\bar{U}_l^{0}$ is the learned initializer}
          \State $\bar{U}_l^{c-1}\leftarrow U_l^{c-1}$
          \State $\mathcal{R}_l^{c}\leftarrow\mathrm{Retrieve}\!\big(U_l^{c-1},\mathcal{E}_l^{c-1}\big)$
          \State $(\mathcal{B}_l^{c},\mathcal{H}_l^{c})\leftarrow\mathrm{Assemble}\!\big(\mathcal{R}_l^{c},\mathcal{H}_l^{c-1},\mathcal{W}_l^{c-1}\big)$
          \State $\mathcal{E}_l^{c}\leftarrow\mathrm{UpdateEntities}\!\big(\mathcal{E}_l^{c-1},U_l^{c-1},\mathcal{W}_l^{c-1}\big)$
        \EndFor
        \ForAll{shallow and deep layers $l$}
          \State $\mathcal{B}_l^{c}\leftarrow\mathrm{UpdateRawWindow}\!\big(\mathcal{B}_l^{c-1},\bar{K}_l^{c-1},\bar{V}_l^{c-1}\big)$
        \EndFor
      \Else
        \State $\mathcal{B}_l^{1}\leftarrow\emptyset$, $\mathcal{H}_l^{1}\leftarrow\emptyset$ for all layers $l$ \Comment{no historical memory before the first chunk}
      \EndIf
      \State $V^{c}\leftarrow\mathrm{Denoise}\!\big(p_c,\,\{\mathcal{C}_l^{c}\}_l,\,\{\mathcal{B}_l^{c}\}_l,\,\{\mathcal{C}_{l}^{\mathrm{text},c}\}_l\big)$
      \State obtain $(\bar{K}_l^{c},\bar{V}_l^{c})$ for all $l$ via $\mathrm{CleanReencode}(V^{c})$
      \State update the local caches and stage $\bar{K}_l^{c},\bar{V}_l^{c}$ as pending memory \Comment{memory from chunk $c$ is read only from chunk $c+1$ onward}
    \EndFor
  \end{algorithmic}
\end{algorithm}

\section{Training and Implementation}
\label{app:impl}

\subsection{Streaming Rollout and Supervised Region}
\label{app:training_rollout}

Memory is trained on the model's own rollout rather than on ground-truth video.
Each iteration advances a virtual sequence of up to $240$ latent frames: the generator synthesizes a chunk,
writes the completed chunk's clean state into memory, and conditions subsequent chunks on that memory,
so slot routing and entity retrieval are exercised under the same self-generated distribution as at inference.
A supervised chunk spans $21$ latent frames, of which at least $18$ are newly generated;
the remaining leading frames are copied from the previous chunk as overlap and serve as context only.
Only newly generated frames receive gradients: both the distillation objective and the fake-score objective
are averaged over a binary frame mask that excludes the overlap, and overlap frames are detached.
Before scoring, the first frame of the window is decoded and re-encoded through the VAE to reduce the train--inference
discrepancy of the conditioning frame.
Prompt switches are sampled inside the rollout from the frame set $\{21,39,\dots,201\}$,
so memory is trained on early, middle, and late transitions within a sequence.

\subsection{Training Objectives}
\label{app:training_details}

Three networks participate. The generator $G_\theta$ and the fake-score network $C_\phi$ are both 1.3B variants,
whereas the real-score teacher $R$ is a frozen 14B model.
$C_\phi$ is updated every optimization step and $G_\theta$ every fifth step,
so the two alternate on the same self-generated sequence.

\noindent\textbf{Generator distribution matching.}
The generator uses a flow-matching parameterization:
for a clean latent $x$, noise $\varepsilon$, and noise level $\sigma_t$,
$x_t=(1-\sigma_t)x+\sigma_t\varepsilon$, $v^\star=\varepsilon-x$,
and a prediction $\hat v$ is converted to $\hat x_0=x_t-\sigma_t\hat v$.
Both score networks predict $x_0$ at the same $x_t$, with classifier-free guidance applied to the real score
($w_{\mathrm{real}}{=}3.0$) and disabled for the fake score ($w_{\mathrm{fake}}{=}0$).
The distribution-matching direction is
\begin{equation}
g=\frac{\hat x_0^{\mathrm{fake}}-\hat x_0^{\mathrm{real}}}
{\operatorname{mean}\lvert x-\hat x_0^{\mathrm{real}}\rvert},
\end{equation}
followed by non-finite-value replacement, and the generator loss is
\begin{equation}
\mathcal L_{\mathrm{DMD}}=\tfrac12\,\mathbb E\big[(x-\operatorname{sg}(x-g))^2\big],
\end{equation}
where the target is stop-gradient.
The expectation is a masked mean over newly generated frames.
Noise levels are integers sampled uniformly from $[0,1000)$, passed through the shift transformation with factor $5.0$,
and clamped to $[20,980]$.

\noindent\textbf{Fake-score denoising.}
The fake score is trained on detached rollouts by regressing the flow target of the generated sample,
$\mathcal L_{\mathrm{critic}}=\mathbb E\big[(\hat v_\phi-(\varepsilon-x))^2\big]$
with $\hat v_\phi=(x_t-\hat x_0^{\mathrm{fake}})/\sigma_t$,
again averaged over newly generated frames only.
The critic objective contains no memory terms.

\noindent\textbf{Slot regularization.}
Let $U$ and $A$ be the current chunk's slot descriptors and soft assignments,
and let $\bar U$ denote the previous chunk's clean slots, which are detached.
Three terms supervise the compressor; each is accumulated once per differentiable forward
and averaged over the forwards within an optimization step.
(i) Temporal consistency treats the same-index slots of consecutive chunks as positives
and all $B{\cdot}S$ slots in the batch as negatives~\citep{manasyan2025temporally}, with $\ell_2$-normalized slots and temperature $\tau_c{=}0.1$:
\begin{equation}
\mathcal L_{\mathrm{con}}=-\frac{1}{BS}\sum_{i=1}^{BS}
\log\frac{\exp\big(\mathrm{sim}(\bar u_i,u_i)/\tau_c\big)}
{\sum_{j=1}^{BS}\exp\big(\mathrm{sim}(\bar u_i,u_j)/\tau_c\big)}.
\end{equation}
(ii) Feature fidelity decodes each slot with $\psi_l$ and reconstructs tokens under a straight-through hard assignment
(hard forward, soft backward); the per-token loss is $1-\cos(\hat x_{b,t},\operatorname{sg}(\bar K_{b,t}))$,
weighted by the top-1 assignment confidence clamped at $0.1$ and by an area weight that compensates small-token slots
(exponent $0.25$, normalized to mean one, capped at five).
(iii) Selection distillation matches the token-selection distribution to the tokens the chunk actually uses:
the teacher attends the chunk's own clean keys and values with its clean queries and weights attention mass by value magnitude,
the student is a softmax over the selection score ($1.5\cos+0.3\log A$) with temperature $0.2$,
and the loss is their cross-entropy over tokens, averaged over batch and slots.
The teacher is computed without gradients.

The generator objective is therefore
\begin{equation}
\mathcal L_{\mathrm{gen}}=\mathcal L_{\mathrm{DMD}}+
\lambda_{\mathrm{slot}}\big(\mathcal L_{\mathrm{con}}
+0.01\,\mathcal L_{\mathrm{rec}}
+0.01\,\mathcal L_{\mathrm{dist}}\big),
\qquad \lambda_{\mathrm{slot}}=0.2.
\end{equation}
Table~\ref{tab:appendix_losses} summarizes all losses, their inputs, their active stage, and their weights.

\begin{table}[t]
\centering
\caption{\textbf{Loss terms and their roles.} ``Generator updates'' means the steps on which $G_\theta$ is optimized.}
\label{tab:appendix_losses}
\setlength{\tabcolsep}{4pt}
\renewcommand{\arraystretch}{1.08}
\resizebox{\linewidth}{!}{
\begin{tabular}{llllc}
\toprule
\textbf{Loss} & \textbf{Target behavior} & \textbf{Inputs} & \textbf{Active} & \textbf{Weight} \\
\midrule
$\mathcal L_{\mathrm{DMD}}$ & match the generator distribution to the teacher & generated latents, noise level, prompt & generator updates & $1.0$ \\
$\mathcal L_{\mathrm{critic}}$ & fit the fake score to generated samples & detached rollouts, flow target & every step & $1.0$ \\
$\mathcal L_{\mathrm{con}}$ & keep slot identity across chunks & consecutive-chunk slot descriptors & all compressor updates & $\lambda_{\mathrm{slot}}$ \\
$\mathcal L_{\mathrm{rec}}$ & ground slots in transformer features & clean keys, decoded prototypes, assignments & unfrozen-grouper updates & $0.01\lambda_{\mathrm{slot}}$ \\
$\mathcal L_{\mathrm{dist}}$ & align selection with actual token usage & clean queries/keys/values, selection scores & unfrozen-grouper updates & $0.01\lambda_{\mathrm{slot}}$ \\
\bottomrule
\end{tabular}}
\end{table}

\subsection{Frozen Modules and Optimization Schedule}
\label{app:training_schedule}

The teacher, the text encoder, and the VAE are frozen.
The 1.3B generator and fake-score backbones are also frozen and adapted only through LoRA
(rank $256$, $\alpha{=}256$, dropout $0$, bf16) applied to the attention blocks;
the memory compressor is trained in full.
Training proceeds in two stages over $3000$ updates.
Stage-I (the first $1000$ updates) freezes the slot grouper, the initializer, and the dynamics predictor,
so the input/output adapters and diffusion-side parameters adapt to a fixed slot vocabulary;
only the contrastive term is accumulated, while reconstruction and selection distillation are disabled.
Stage-II (the remaining $2000$ updates) unfreezes these modules and resumes from the Stage-I checkpoint,
enabling end-to-end slot optimization with all three memory losses.
The stage boundary is implemented by a configuration flag plus an early-stopping check at update $1000$.
Throughout, ``freezing the slot grouper'' refers exactly to these three modules; the adapters are never frozen.
The entity bank has no trainable parameters: it is a state machine updated without gradients,
and it is therefore not optimized by any loss.
EMA is disabled in the LoRA configuration, and the two backbones are optimized alternately as described above.

\subsection{Training Recipe}
\label{app:training}

The reported model starts from a released causal base checkpoint and initializes the slot modules
from the slot-attention checkpoint of \citet{manasyan2025temporally}, pretrained on real videos.
Training takes $3000$ updates on 8 GPUs with FSDP, split into the $1000$-update frozen-grouper stage
and the $2000$-update end-to-end stage described above.
The generator uses AdamW with learning rate \(1\times10^{-5}\) and the critic with \(2\times10^{-6}\)
(\(\beta_1{=}0\), \(\beta_2{=}0.999\), weight decay \(0.01\)), batch size \(1\), and gradient accumulation \(1\).
Checkpoints are written every $50$ updates (keeping the most recent three) and validation videos every $500$ updates.

The streaming rollout uses chunks of $21$ latent frames with a maximum sequence length of $240$ frames,
at least $18$ newly generated frames per chunk, and $3$ latent frames per generation block.
Each block is denoised with the step schedule \(\{1000,750,500,250\}\) and a clean re-encoding pass at zero context noise.
Prompt switches are drawn uniformly from frame indices \(\{21,39,\dots,201\}\), covering early, middle, and late transitions.
The teacher is a frozen Wan2.1-T2V-14B model~\citep{wan2025wan}, the fake score is a 1.3B model, and the generator is a causal 1.3B model;
guidance is $3.0$ on the real score and disabled on the fake score,
and the timestep shift is $5.0$ with denoising loss type \texttt{flow}.

\section{Additional Quantitative Results}
\label{app:quantitative}

This section extends the main quantitative evaluation along the two axes that \Secref{sec:exp_main} does not cover:
horizon scaling beyond 60\,s and end-to-end efficiency.

\subsection{Long-Horizon Scaling}
\label{app:longhorizon}

\begin{table*}[t]
\centering
\caption{\textbf{Quantitative comparison on long video generation.} We report VBench-Long~\citep{huang2025vbench++} metrics. Within each base model group, best results are bold and second best are underlined. Baseline results are taken from MemRoPE~\citep{kim2026memrope} and LongLive-RAG~\citep{hu2026longliverag}. ``--'' indicates unavailable results and is counted as the last rank when computing Avg. Rank.}
\label{tab:main}
\resizebox{\textwidth}{!}{%
\begin{tabular}{l cccccc !{\vrule width 0.8pt} c}
\toprule
Method & \makecell{Subject\\Consistency} $\uparrow$ & \makecell{Background\\Consistency} $\uparrow$ & \makecell{Motion\\Smoothness} $\uparrow$ & \makecell{Dynamic\\Degree} $\uparrow$ & \makecell{Aesthetic\\Quality} $\uparrow$ & \makecell{Imaging\\Quality} $\uparrow$ & \makecell{Avg.\\Rank} $\downarrow$ \\
\midrule
\multicolumn{8}{c}{\textit{120 seconds}} \\
\midrule
\multicolumn{8}{l}{\textit{Results on LongLive}~\citep{yang2025longlive}} \\
LongLive~\citep{yang2025longlive}          & 96.93 & 95.64 & 98.58 & 47.12 & 57.90 & 66.95 & 4.08 \\
Deep Forcing~\citep{yi2025deep}            & 97.17 & 95.72 & 98.57 & 46.03 & 56.98 & 66.03 & 4.58 \\
Infinity-RoPE~\citep{yesiltepe2025infinity}& 96.81 & 95.65 & 98.48 & \underline{53.59} & 56.73 & 66.19 & 5.00 \\
MemRoPE~\citep{kim2026memrope}             & \underline{97.54} & \textbf{96.29} & \textbf{98.73} & -- & \underline{59.25} & \textbf{69.43} & \textbf{2.17} \\
LongLive-RAG~\citep{hu2026longliverag}     & 97.22 & \underline{95.88} & \underline{98.62} & 50.25 & 57.57 & 66.95 & 2.92 \\
\textbf{\ours{}}                           & \textbf{97.59} & 95.72 & 98.56 & \textbf{78.26} & \textbf{62.27} & \underline{68.92} & \underline{2.25} \\
\midrule
\multicolumn{8}{c}{\textit{240 seconds}} \\
\midrule
\multicolumn{8}{l}{\textit{Results on LongLive}~\citep{yang2025longlive}} \\
LongLive~\citep{yang2025longlive}          & 97.02 & 95.80 & 98.52 & 33.21 & 56.70 & 66.90 & 3.83 \\
Deep Forcing~\citep{yi2025deep}            & 96.65 & 96.03 & 98.50 & \underline{39.17} & 57.75 & 66.48 & 3.50 \\
Infinity-RoPE~\citep{yesiltepe2025infinity}& 97.11 & \underline{96.34} & \textbf{98.99} & -- & 57.47 & 63.38 & 3.33 \\
MemRoPE~\citep{kim2026memrope}             & \underline{97.37} & \textbf{96.39} & \underline{98.59} & 38.25 & \underline{58.90} & \textbf{68.93} & \textbf{1.83} \\
\textbf{\ours{}}                           & \textbf{97.67} & 95.77 & 97.71 & \textbf{77.96} & \textbf{61.97} & \underline{68.57} & \underline{2.50} \\
\bottomrule
\end{tabular}%
}
\end{table*}

The 30\,s and 60\,s results in \Secref{sec:exp_main} show that memory matters more as the rollout lengthens,
but they cannot distinguish whether the advantage of entity-addressed memory keeps accumulating with horizon or saturates while
temporal memories continue to decay. Table~\ref{tab:main} separates these possibilities at 120\,s and 240\,s, and the answer is
the second: our scores are essentially horizon-invariant whereas every baseline that reports the diagnostic dimension degrades,
so what grows with horizon is the separation between methods, not our absolute quality. This is the distinction the long-horizon setting is designed to isolate.

The six metrics divide into two families that respond oppositely to what a memory stores.
Subject Consistency, Dynamic Degree, and Aesthetic Quality measure whether appearance and motion survive;
Background Consistency and Motion Smoothness reward re-using or stabilizing scene content, and Motion Smoothness is
highest for video that changes little. \ours{} leads the first family at both horizons (Subject 97.59/97.67,
Dynamic 78.26/77.96, Aesthetic 62.27/61.97) but ranks fifth in the second at 240\,s, with Background Consistency 95.77
and Motion Smoothness 97.71. The second-family deficit is small in absolute terms---0.62 behind the best background score and
0.79--1.28 behind the smoothest method---whereas the motion gap is 38.8--44.7 points.
This asymmetry is the signature of the design: a bank that stores entity-bearing K/V tokens instead of replaying recent scene
content leaves the scene free to move, so it gives up some smoothness and background stability in exchange for motion.
The same trade-off is already visible at 60\,s, where the replay-based baselines collapse in Dynamic Degree
(LongLive 28.30, MemFlow 26.05) while \ours{} reaches 81.38; at 240\,s it becomes the dominant axis of separation.

The scaling pattern shows that this behaviour is not a transient of the 30--60\,s regime.
Across horizons our Dynamic Degree is 79.30 at 30\,s, 81.38 at 60\,s, 78.26 at 120\,s, and 77.96 at 240\,s---a spread of
3.4 points over an eightfold increase in length (the 60\,s protocol is interactive while the others are single-prompt)---and our Subject Consistency and Aesthetic Quality change by at most 0.35
points between the two long horizons. The baselines move in the opposite direction: between 120\,s and 240\,s LongLive loses
13.9 Dynamic points and Deep Forcing 6.9, and the only Dynamic value reported by MemRoPE at 240\,s is 38.25.
Two consequences follow. First, appearance metrics saturate (every method is at or above 96.6 on Subject Consistency),
so they stop discriminating and the informative signal moves to the motion dimensions. Second, a fixed bank of 42 groups per semantic layer already supports 240\,s without the decay observed in the baselines, which is the long-horizon form of the paper's capacity argument: what binds is what the memory stores, not how long the video is, because entity addresses are independent of temporal distance.

The average-rank column must be read against this split, since it weights both families equally.
At 240\,s \ours{} is first on Subject Consistency, Dynamic Degree, and Aesthetic Quality but fifth on Background Consistency
and Motion Smoothness, which yields a rank of 2.50; MemRoPE attains 1.83 by leading Background and Imaging Quality while
not reporting Dynamic Degree at 120\,s and reaching only 38.25 at 240\,s. A memory that keeps a stable, slowly changing scene
can therefore improve an aggregate rank without preserving the motion that the dynamic metrics are designed to capture.
We accordingly treat the per-dimension pattern, rather than the aggregate rank, as the evidence about memory quality.

\subsection{Efficiency and Memory Overhead}
\label{app:efficiency}

\begin{table}[t]
    \centering
    \caption{Efficiency comparison for interactive long video generation on a single NVIDIA H200 GPU.
    We report the average latency, average peak GPU memory usage, and diffusion-only throughput.}
    \label{tab:efficiency}
    \resizebox{\linewidth}{!}{%
    \begin{tabular}{lccc}
        \toprule
        \textbf{Method} &
        \textbf{Latency (s)$\downarrow$} &
        \textbf{Peak VRAM (GB)$\downarrow$} &
        \textbf{Diffusion FPS$\uparrow$} \\
        \midrule
        LongLive~\citep{yang2025longlive}        & 73.040 $\pm$ 0.398 & 33.24 & 18.3 \\
        MemFlow~\citep{ji2025memflow}         & 81.423 $\pm$ 0.000 & 31.47 & 16.2 \\
        Anchor Forcing~\citep{yang2026anchorforcing}  & 88.269 $\pm$ 0.354 & 34.15 & 14.6 \\
        \textbf{\ours{}} & 83.005 $\pm$ 0.281 & 36.82 & 14.9 \\
        \bottomrule
    \end{tabular}}
\end{table}

Table~\ref{tab:efficiency} measures the interactive protocol on a single H200 GPU.
\ours{} completes the sequence in 83.0\,s at 14.9 diffusion FPS, which is 13.6\% slower than LongLive (73.0\,s, 18.3 FPS) and 1.9\% slower than MemFlow (81.4\,s, 16.2 FPS), while remaining 6.0\% faster than Anchor Forcing (88.3\,s, 14.6 FPS).
Its peak memory of 36.8\,GB is 3.6, 5.4, and 2.7\,GB above LongLive, MemFlow, and Anchor Forcing, respectively, the largest relative increase (17.0\%) being over MemFlow.
The structured memory therefore adds a bounded overhead: the method's latency stays within the range of the evaluated memory baselines and it remains faster than one of them, at the cost of the largest peak memory among the four measured systems.

\subsection{RoPE and Positional Encoding Analysis}
\label{app:rope}

A remaining question about the entity bank is whether its identity key is confounded by absolute position.
Keys are rotated by RoPE before they are cached, and the identity pathway reads the same cached tensors, so a slot descriptor
summarizes content whose phase encodes \emph{when} it occurred as well as \emph{what} it is. Because association is decided by
the cosine similarity between slot descriptors, two occurrences of the same entity that differ only in temporal phase can be
separated even when their content is identical. The concern is therefore one of matching rather than attention: an entangled
\emph{who}--\emph{when} representation could weaken long-range association and complicate the matching gate.

We examined this question with three configurations that share the attention payload (RoPE-rotated K/V) and differ only
in how the identity key is obtained: grouping slots from rotated keys and using the resulting descriptor, which is the
configuration adopted in this work; grouping from rotated keys but pooling pre-rotation keys for identity; and deriving both
grouping and identity from pre-rotation keys. Table~\ref{tab:appendix_rope} reports the comparison on the entity-bank variant
without selection distillation.

The interventions change the consistency dimensions in the direction predicted by the confound hypothesis, but only
locally. Deriving identity from pre-rotation slots yields the highest Subject Consistency (93.31), and pooling pre-rotation keys
yields the lowest alignment drift (6.98), confirming that the absolute phase affects long-range matching. Both interventions,
however, cost more than they recover: the pre-rotation-slot variant gives up 3.93 points of Dynamic Degree and 0.51 points of
VBench Quality relative to the adopted configuration, and the pooled-key variant gives up 0.19 points of Quality for its
0.86-point reduction in drift. Across the three configurations the adopted design remains the strongest on average alignment
(66.63), Dynamic Degree (92.96), and overall quality (81.50), while the alternatives lead only on individual dimensions.

We therefore retain the coupled form, in which the identity key is computed from the same rotated representation in
which the recalled snapshot is consumed by attention. This keeps association and retrieval in one space: a match then means that
the content compared is the content that will be attended to. Decoupling identity from the payload moves the matcher into a space the generator does not attend in, which is consistent with the alternatives' local gains not translating into aggregate quality; deriving grouping from pre-rotation keys additionally changes slot formation, and this variant loses the most motion. The phase is not pure nuisance either, since the relative phase between two keys encodes their temporal proximity, a
useful cue for associating the same entity across nearby chunks and for separating similar-looking entities. 

\begin{table}[t]
\centering
\caption{\textbf{Effect of the positional representation used for entity identity.} All configurations share the same attention payload (RoPE-rotated K/V) and differ only in how the identity key is obtained. Text alignment is averaged over the six 10\,s segments and drift is the drop from 0--10\,s to 50--60\,s. Results use the entity-bank variant without selection distillation under the 60\,s interactive protocol; the configuration adopted in this work is indicated and the best value per metric is shown in \textbf{bold}.}
\label{tab:appendix_rope}
\setlength{\tabcolsep}{4pt}
\renewcommand{\arraystretch}{1.08}
\resizebox{\linewidth}{!}{
\begin{tabular}{llccccc}
\toprule
\textbf{Identity key} & \textbf{Slot grouping} & \makecell{Text\\alignment $\uparrow$} & \makecell{Alignment\\drift $\downarrow$} & \makecell{Subject\\consistency $\uparrow$} & \makecell{Dynamic\\degree $\uparrow$} & \makecell{VBench\\quality $\uparrow$} \\
\midrule
Rotated-key slot (adopted) & Rotated keys & \textbf{66.63} & 7.84 & 93.07 & \textbf{92.96} & \textbf{81.50} \\
Pooled pre-rotation key & Rotated keys & 66.59 & \textbf{6.98} & 92.89 & 91.42 & 81.31 \\
Pre-rotation slot & Pre-rotation keys & 65.85 & 7.33 & \textbf{93.31} & 89.03 & 80.99 \\
\bottomrule
\end{tabular}}
\end{table}

\section{Human Evaluation Protocol}
\label{app:human_evaluation}

We conducted a blinded six-way human evaluation to assess the overall perceptual quality of long-form interactive videos generated by \ours{} and five baseline methods. The study was designed to complement the automatic metrics by jointly evaluating perceptual properties that are difficult to capture with a single numerical measure, including long-term identity preservation, action completion, scene consistency, temporal stability, and object interaction.

\subsection{Design and Stimuli}
\label{app:human_stimuli}

The evaluation used the 100 interactive scripts of the main benchmark. For each script we generated videos with six methods---\ours{}, Anchor-Forcing~\citep{yang2026anchorforcing}, IAMFlow~\citep{iamflow}, Infinity-RoPE~\citep{yesiltepe2025infinity}, LongLive~\citep{yang2025longlive}, and MemFlow~\citep{ji2025memflow}---conditioned on exactly the same prompt sequence, and repeated generation with three seeds, giving $100\times3=300$ six-way comparison groups. Each group therefore contained one output per method under matched textual conditions, so differences within a group reflect the generation methods rather than the conditions.

Stimuli were normalized to a common spatial resolution, frame rate, encoding format, bitrate, and duration, and audio tracks were removed. Method names, filenames, watermarks, and other identifying metadata were stripped, and no method-specific post-processing was applied.

Each trial presented the six videos of one group anonymously in a $2\times3$ grid labelled A--F, with the method-to-position mapping randomized independently per trial and approximately balanced across the study to reduce position bias. All videos used identical player dimensions and playback settings; participants could replay them but had to watch each video in full before responding, and the interface exposed no method-specific information.

\subsection{Procedure and Evaluation Criteria}
\label{app:human_procedure}

We recruited 20 adult participants. Participation was voluntary, participants were informed of the procedure, the task, and their right to withdraw, and no sensitive personal information was collected.

Each participant evaluated 15 groups, giving $20\times15=300$ valid overall-preference judgments; each script--seed group was judged once, and both the assignment of groups to participants and the presentation order were randomized. After watching all six videos of a trial, participants selected the single best overall video, jointly considering five criteria: identity preservation (consistent appearance, clothing, colour, and role of recurring characters and objects); action completion (prompted actions visibly completed within the corresponding segments); scene consistency (background, layout, and persistent elements coherent across transitions unless the script requests a change); temporal stability (no flicker, abrupt appearance changes, deformation, or inconsistent transitions); and object interaction (plausible contact, spatial relations, and action consequences). Participants were explicitly told that stability is not the absence of motion and that a nearly static video should not be preferred for containing little variation; stability was to be judged together with whether the prompted actions, motion, and scene changes occurred. Each trial yielded a single overall-preference label rather than separate ratings per criterion.

\subsection{Validity and Quality Control}
\label{app:human_quality_control}

Trials that could not be compared reliably, for example because of a playback failure, could be marked invalid; such responses were never treated as preferences, and the group was reassigned and re-evaluated until one valid judgment per script--seed group was obtained. Participant-level or trial-level data were excluded only under pre-specified criteria: an incomplete session, a recorded viewing duration indicating that the videos were not watched in full, a failed attention check, or severe inconsistency on repeated quality-control trials. Repeated trials served only to detect substantial inconsistency and were not part of the 300 preference labels. No response was excluded for disagreeing with the expected ranking or the authors' hypotheses. Because each group received one valid label, the study provides broad coverage across conditions rather than an estimate of inter-rater agreement.

\subsection{Aggregation and Results}
\label{app:human_results}

Each method's preference rate is \(P_m=N_m/N\), where \(N_m\) is the number of valid trials in which the method was selected and \(N=300\); every valid trial selected exactly one method, so \(\sum_m N_m=300\). These rates are aggregate preference frequencies, not absolute quality scores.

\begin{table}[t]
    \centering
    \caption{
        Results of the blinded six-way human evaluation.
        Each of the 300 script--seed groups contributed one valid
        overall-preference label. Participants selected one video after
        jointly considering identity preservation, action completion,
        scene consistency, temporal stability, and object interaction.
    }
    \label{tab:human_preference}
    \setlength{\tabcolsep}{8pt}
    \begin{tabular}{lrr}
        \toprule
        \textbf{Method}
        & \textbf{Selected Count}
        & \textbf{Preference Rate} \\
        \midrule
        \ours{}       & \textbf{81} & \textbf{27.0\%} \\
        Anchor-Forcing~\citep{yang2026anchorforcing}   & 62 & 20.7\% \\
        LongLive~\citep{yang2025longlive}         & 52 & 17.3\% \\
        Infinity-RoPE~\citep{yesiltepe2025infinity}    & 43 & 14.3\% \\
        IAMFlow~\citep{iamflow}          & 37 & 12.3\% \\
        MemFlow~\citep{ji2025memflow}          & 25 & 8.3\%  \\
        \midrule
        Total            & 300         & 100.0\%          \\
        \bottomrule
    \end{tabular}
\end{table}

Table~\ref{tab:human_preference} reports the results. \ours{} was selected in 81 of the 300 groups (\(27.0\%\)), ahead of Anchor-Forcing (62; \(20.7\%\)), LongLive (52; \(17.3\%\)), Infinity-RoPE (43; \(14.3\%\)), IAMFlow (37; \(12.3\%\)), and MemFlow (25; \(8.3\%\)). \ours{} was therefore the most frequently preferred method, by 6.3 percentage points over the second-ranked Anchor-Forcing and 9.7 points over LongLive.

\section{More Qualitative Evidence}
\label{app:qualitative}

\begin{figure*}[t]
\centering
\includegraphics[width=\textwidth]{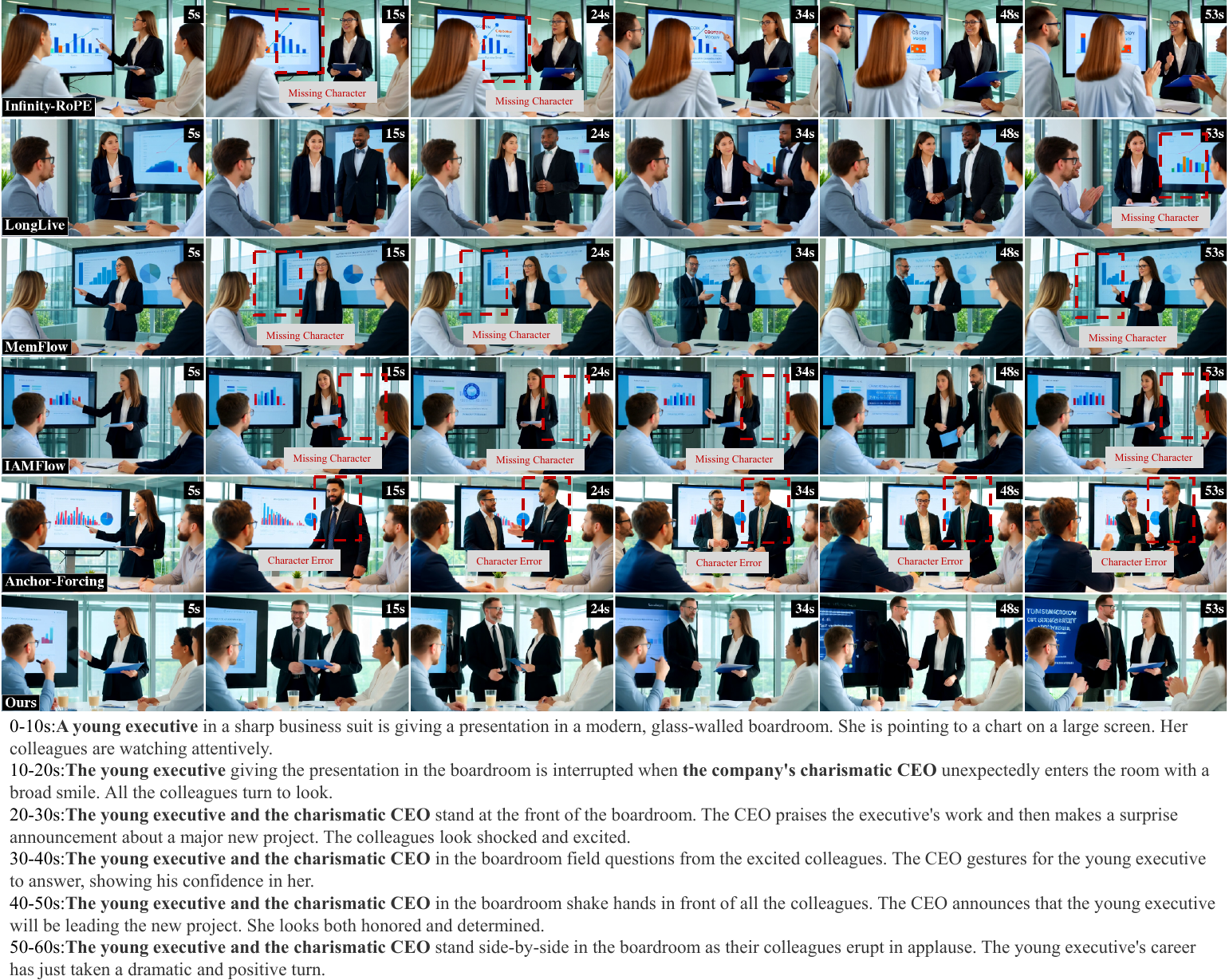}
\caption{\textbf{Extended comparison on a dialogue-driven script with a cast change.}
The narrative moves from a presentation by a young executive to a joint announcement with the CEO, who enters at the
second prompt, and ends with the two characters sharing the stage. Each row shows one method over the six segments.
The case tests whether the original subject is retained while a new character is introduced and the interaction
remains coherent; the cast change is the first point at which a temporal memory can lose the initial subject.}
\label{fig:qual_boardroom}
\end{figure*}

\begin{figure*}[t]
\centering
\includegraphics[width=\textwidth]{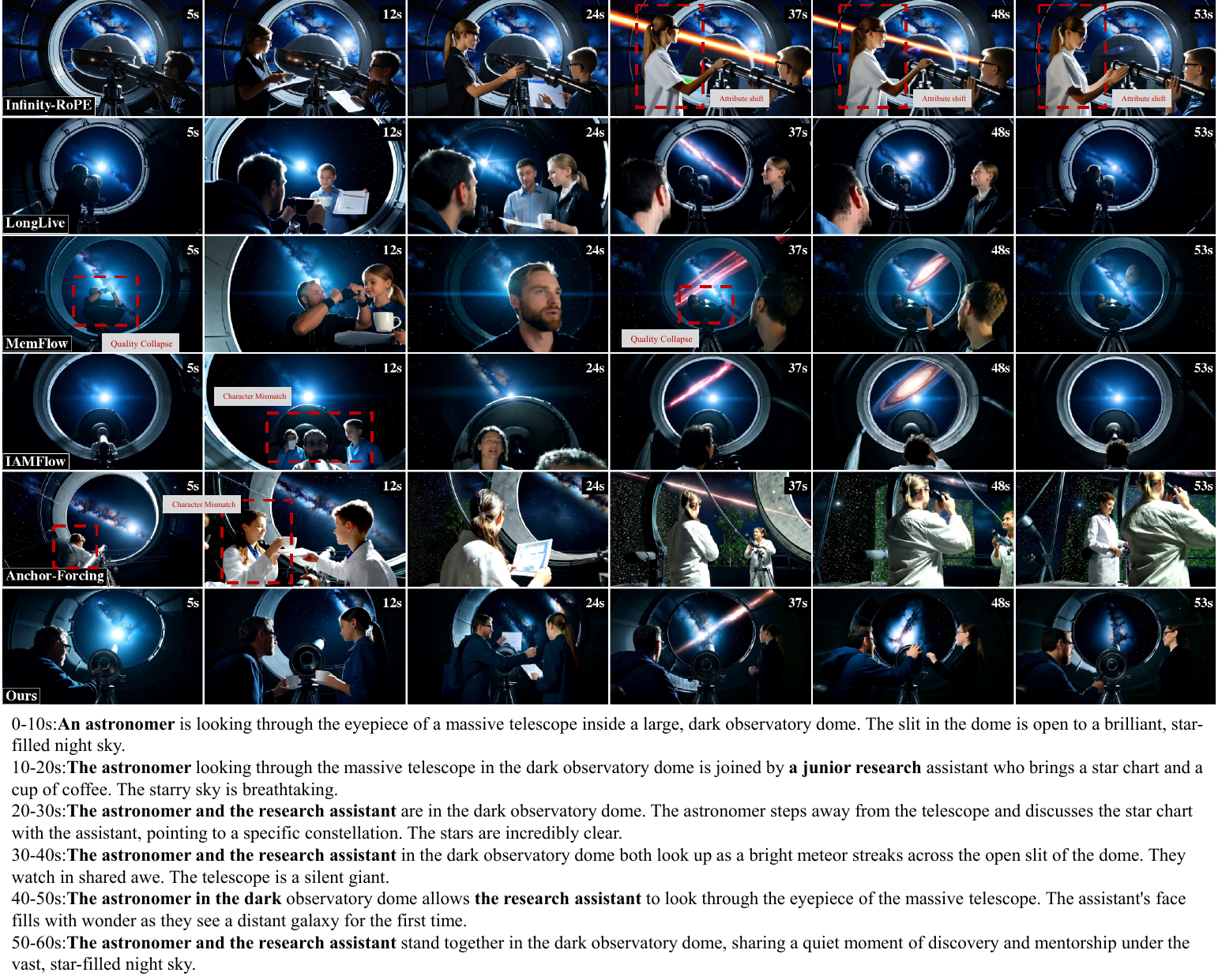}
\caption{\textbf{Extended comparison on a two-character script with a transient event.}
An astronomer is joined by a research assistant in an observatory, where a bright meteor briefly crosses the sky.
The case combines a persistent two-character cast, a low-light background, and a short event that must not dominate
the memory. Labels mark the failure modes observed in the compared outputs, including attribute shift, quality
collapse, and character mismatch.}
\label{fig:qual_observatory}
\end{figure*}

\begin{figure*}[t]
\centering
\includegraphics[width=\textwidth]{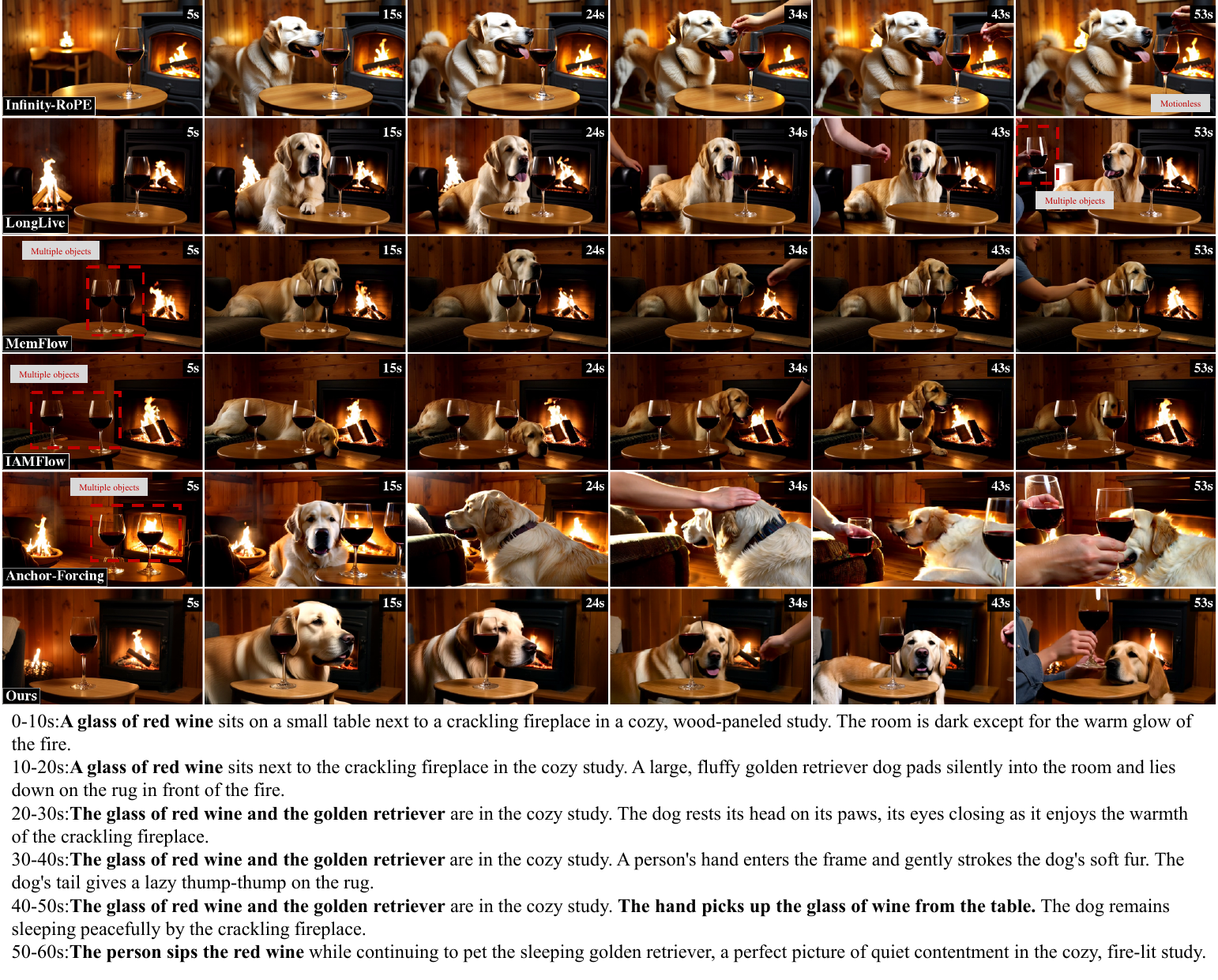}
\caption{\textbf{Extended comparison on a low-motion multi-object script.}
A glass of wine and a sleeping dog share a dimly lit study, and a hand enters the frame to stroke the dog and lift
the glass. The scene deliberately contains little motion but several objects whose states change over time.
Labels mark motionless output and spurious additional objects in the compared outputs, which is the failure mode
that consistency-oriented metrics alone do not penalize.}
\label{fig:qual_study}
\end{figure*}

This section provides the extended qualitative evidence behind the claims of \Secref{sec:exp_main}:
identity persistence across prompt switches, adherence to the changing instruction, motion, and visual quality.
The comparisons follow the protocol and baselines of the quantitative interactive evaluation~\citep{yang2025longlive,ji2025memflow,yesiltepe2025infinity,iamflow,yang2026anchorforcing}---six prompts of
$10$\,s and keyframes sampled at matched positions within each segment---so that all methods are examined at the
same narrative moments. As shown in \Figref{fig:qual_boardroom}, \Figref{fig:qual_observatory}, and \Figref{fig:qual_study}, three scripts are included to cover complementary demands on memory rather than to maximize
the margin: a dialogue-driven scene whose cast changes at a switch, a two-character scene with a brief transient
event, and a low-motion scene with several objects.

\begin{figure*}[t]
\centering
\includegraphics[width=\textwidth]{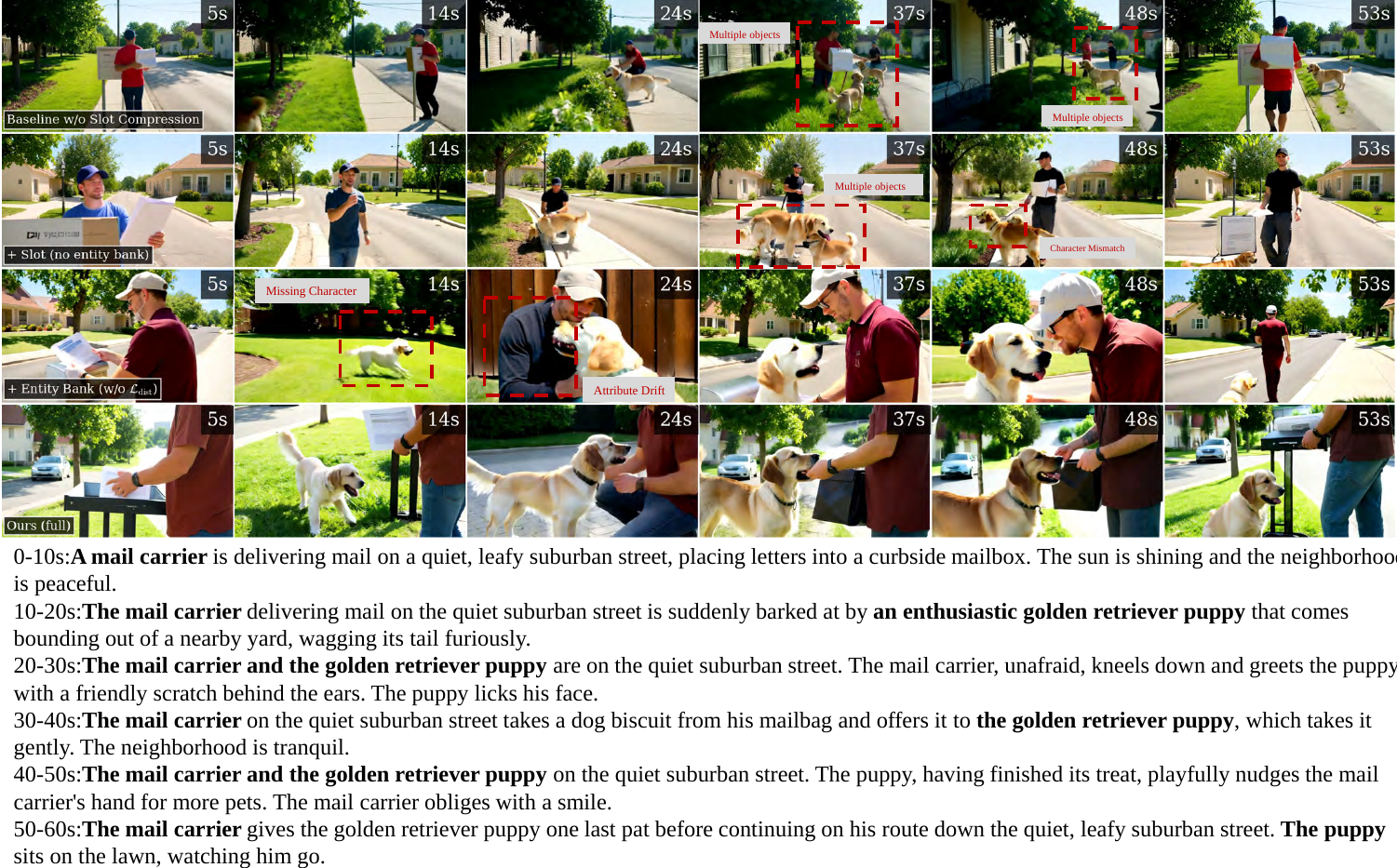}
\caption{\textbf{Visualizing the core mechanism ablation on a two-entity script.}
Rows correspond to the variants of Table~\ref{tab:core_ablation} and columns to the six prompt segments;
labels mark the failure modes observed in the individual outputs, including multiple objects, character mismatch,
attribute drift, and a missing character.}
\label{fig:qual_core_ablation}
\end{figure*}

\noindent\textbf{What the failures reveal.}
Figure~\ref{fig:qual_core_ablation} makes the numerical pattern of the core ablation concrete.
Keeping history as untyped tokens allows the same visual evidence to be stored without an object-level owner, so the
generator can reproduce an entity twice or, once the original subject has left the local window, drop it altogether;
this is the regime in which multiple objects and a missing character appear. Grouping tokens into slots fixes the
object count while the entities are still recent, but grouping alone does not decide that a slot in one chunk denotes
the same object as a slot in another. When that correspondence is absent, the retrieved content can be semantically
plausible while belonging to the wrong state, which surfaces as character mismatch and attribute drift.
The entity bank is what turns a per-chunk decomposition into a persistent correspondence, so the visual difference
between the ablated variants is not a difference in how much memory is kept but in what the kept states represent.

The distribution of the annotations is equally informative: they are sparse and segment-local rather than present in
every frame, which is why an average over the whole sequence dilutes them while metrics restricted to the tail of the
rollout separate the variants. This is the qualitative counterpart of the pattern in Table~\ref{tab:core_ablation},
where the average alignment barely moves but tail alignment and text drift change sharply: what the variants lose is
not general visual quality but the ability to recall an entity correctly once recency can no longer substitute for
memory.

Taken together, the comparisons show where the difference documented in \Secref{sec:exp_main} becomes
visible: the compared methods retain the set or the layout of a scene while losing the identity of the characters
that populate it, or they remain visually stable by changing very little, whereas the proposed method keeps the
cast and the depicted action across the switches.

